\pdfoutput=1

\documentclass[11pt]{article}

\usepackage{EMNLP2022}

\usepackage{times}
\usepackage{latexsym}

\usepackage[T1]{fontenc}

\usepackage[utf8]{inputenc}
\DeclareUnicodeCharacter{0307}{ }

\usepackage{microtype}

\usepackage{inconsolata}

\usepackage{subfigure}
\usepackage{microtype}
\usepackage{bm}
\usepackage{url}
\usepackage{graphicx}
\usepackage{multirow, booktabs}
\usepackage{amsmath}
\usepackage{enumitem}
\usepackage{amsfonts}
\usepackage{makecell}
\usepackage{verbatim}
\usepackage{array}
\usepackage{xcolor}
\usepackage{soul}
\usepackage{longtable}
\usepackage[figuresright]{rotating}

\newcommand{\qinyuan}[1]{\textcolor{teal}{}}
\newcommand{\xiang}[1]{\textcolor{red}{}}
\newcommand{\juan}[1]{\textcolor{blue}{}}
\usepackage{inconsolata}
\usepackage{colortbl}
\usepackage{listings}

\colorlet{punct}{red!60!black}
\definecolor{background}{HTML}{EEEEEE}
\definecolor{delim}{RGB}{20,105,176}
\colorlet{numb}{magenta!60!black}

\lstdefinelanguage{json}{
    basicstyle=\scriptsize\ttfamily,
    numbers=left,
    numberstyle=\scriptsize,
    stepnumber=1,
    numbersep=8pt,
    showstringspaces=false,
    breaklines=true,
    frame=lines,
    backgroundcolor=\color{background},
    literate=
     *{0}{{{\color{numb}0}}}{1}
      {1}{{{\color{numb}1}}}{1}
      {2}{{{\color{numb}2}}}{1}
      {3}{{{\color{numb}3}}}{1}
      {4}{{{\color{numb}4}}}{1}
      {5}{{{\color{numb}5}}}{1}
      {6}{{{\color{numb}6}}}{1}
      {7}{{{\color{numb}7}}}{1}
      {8}{{{\color{numb}8}}}{1}
      {9}{{{\color{numb}9}}}{1}
      {:}{{{\color{punct}{:}}}}{1}
      {,}{{{\color{punct}{,}}}}{1}
      {\{}{{{\color{delim}{\{}}}}{1}
      {\}}{{{\color{delim}{\}}}}}{1}
      {[}{{{\color{delim}{[}}}}{1}
      {]}{{{\color{delim}{]}}}}{1},
}

\title{Eliciting and Understanding Cross-Task Skills \\with Task-Level Mixture-of-Experts}

\author{
Qinyuan Ye \quad Juan Zha \quad Xiang Ren \\
  University of Southern California \\
  \texttt{\{\href{mailto:qinyuany@usc.edu}{qinyuany}, \href{mailto:juanzha@usc.edu}{juanzha}, \href{mailto:xiangren@usc.edu}{xiangren}\}@usc.edu} 
}

\begin{document}
\maketitle
\begin{abstract}
Recent works suggest that transformer models are capable of multi-tasking on diverse NLP tasks and adapting to new tasks efficiently.
However, the potential of these multi-task models may be limited as they use the \textit{same} set of parameters for \textit{all} tasks. 
In contrast, humans tackle tasks in a more flexible way, by making proper presumptions on what skills and knowledge are relevant and executing only the necessary computations.
Inspired by this, we propose to use task-level mixture-of-expert models, which has a collection of transformer layers (\textit{i.e.}, experts) and a router component that chooses from these experts dynamically and flexibly.
We find that these models help improve the average performance gain (ARG) metric by 2.6\% when adapting to unseen tasks in the few-shot setting and by 5.6\% in the zero-shot generalization setting.
Further, we show that the learned routing decisions partly rediscover human categorization of NLP tasks~--~certain experts are strongly associated with extractive tasks, some with classification tasks, and some with tasks requiring world knowledge.\footnote{Our code will be released at \url{https://github.com/INK-USC/CrossTaskMoE}.}
\end{abstract}

\section{Introduction}


Pre-trained transformer models \cite{devlin-etal-2019-bert, liu2019roberta} have demonstrated remarkable capabilities in natural language processing (NLP) in recent years. Moreover, generative transformers can be viewed as a universal model that can be optimized for any language task primed into text-to-text format \cite{t5}. Recently, researchers found that training these transformer models to multi-task on a diverse collection of NLP tasks is beneficial – not only are they better at handling seen tasks \cite{aghajanyan-etal-2021-muppet, aribandi2022ext}, but also at generalizing and adapting to unseen tasks \cite{wei2021finetuned, sanh2022multitask}.

\begin{figure}[t]
    \centering
    \includegraphics[width=0.5\textwidth]{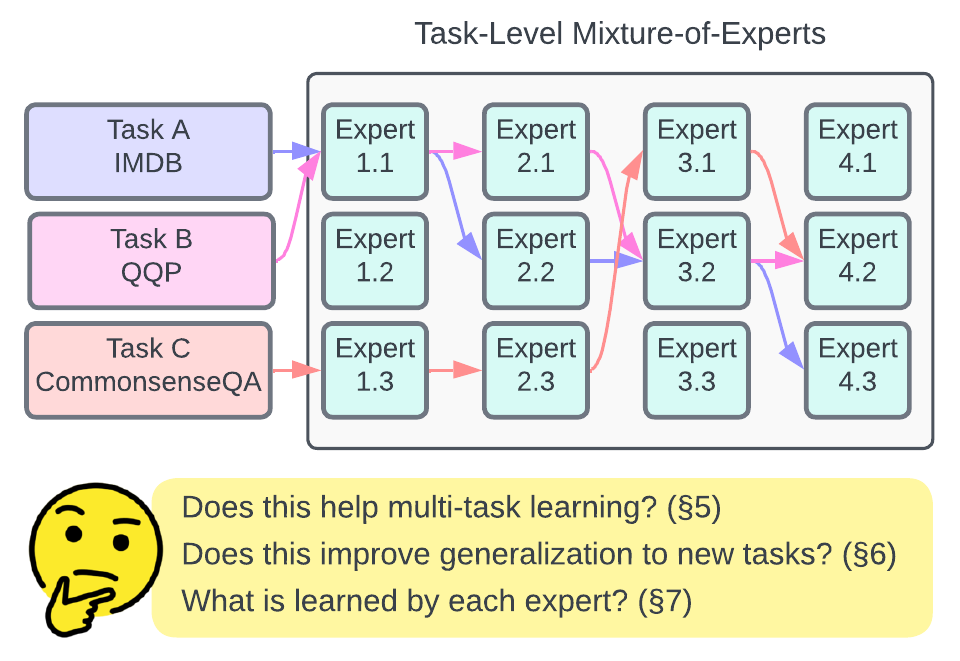}
    \caption{\textbf{Illustration of Task-level Mixture-of-Expert Models.} In this work, we train such models to multi-task on diverse NLP tasks, aiming at modeling skill sharing explicitly and understanding the learned patterns.}
    \label{fig:intro}
\end{figure}

However, little is known about how multi-tasking capabilities and cross-task generalization is achieved, especially that the exact \textit{same} set of weights is applied, and the \textit{same} computation is executed, for very \textit{different} tasks. Humans, on the other hand, do not exhaust their brain capacity for every task at hand. Humans develop skill sets and accumulate knowledge during learning, and can reuse and recompose them when facing a task. 
Inspired by this, we hypothesize that a model that explicitly emulate skill and knowledge sharing may help improve multi-task performance and generalization to new tasks.
A natural fit for this goal would be task-level mixture-of-expert models \cite{JacobsJordanNowlanEtAl91, Kudugunta21Beyond}, where the model computation is conditioned on the task at hand. More specifically, the model contains a collection of experts and a router that chooses from the experts and composes the final model (Fig.~\ref{fig:intro}-\ref{fig:arch}). 

In this paper, we first empirically investigate several key design choices for effectively training task-level mixture-of-experts models (\S\ref{sec:multitask}). 
We further test the model's task-level generalization capabilities by testing it on unseen tasks (\S\ref{sec:few-shot}).
Compared to a multi-task BART-Base \cite{lewis-etal-2020-bart} baseline, our final method leads to an 2.6\% improvement in the average performance gain (ARG) metric when adapting to 18 unseen tasks \cite{ye-etal-2021-crossfit} in the few-shot learning setting. Further, a gain of 5.6\% in ARG is obtained in the zero-shot setting with P3 dataset \cite{sanh2022multitask}.
Lastly, we conduct a detailed analysis quantifying the correlations between the learned routes and the characteristics of tasks (\S\ref{sec:interpret}). 
We find that the routing decisions, though learned purely from multi-tasking \textit{without} prior knowledge, strongly correlate with human understanding of task characteristics, such as the task being a classification task, the task being extractive, or the task requiring world knowledge.

\section{Related Work}
\paragraph{Massive Multi-task Learning.} Multi-task learning \cite{caruana1997multitask} has been continuously explored in NLP and is shown to be beneficial \cite{mccann2018natural, liu-etal-2019-multi}. 
Recently, multi-task learning in NLP is brought to a new scale by using a significantly larger collection of tasks and examples \cite{aghajanyan-etal-2021-muppet, aribandi2022ext, khashabi-etal-2020-unifiedqa, hendryckstest2021}. These work demonstrate that multi-task learning improves the learning of text representation and thus boost the performance of seen tasks.
Moreover, these models also exhibit strong adaptability to unseen tasks, in both few-shot \cite{ye-etal-2021-crossfit} and zero-shot settings \cite{wei2021finetuned, sanh2022multitask, mishra2021cross}. Despite their effectiveness in terms of performance, how a model learns and spontaneously develops language skills during multi-task learning is a relatively under-explored topic. In our work, we try to investigate this question by training task-level MoE models and interpreting them. We additionally discuss contemporary works \cite{ponti2022combining, gupta2022sparsely,asai2022attempt} in Appendix \ref{app:contemporary-works}.

\paragraph{Mixture-of-Experts in NLP.} Mixture-of-experts models \citep{JacobsJordanNowlanEtAl91} divide the problem space into several sub-spaces and allow experts to be specialized in each subspace. Recently this concept is successfully applied to NLP \cite{ShazeerMMDLHD17}, enabling models of billion or even trillion parameter scale \cite{fedus2021switch, Du2021GLaMES, mikel2021efficient, zoph2022designing}. However these applications mainly focus on the \textit{scaling} aspects. Besides, most of them select experts on a per-example or per-token basis. 
In this work we are interested in multi-task learning with per-task gating decisions \cite{rosenbaum2018routing, Kudugunta21Beyond}, and mainly focus on understanding and interpreting task transferability.

\paragraph{Task Transferability in NLP.} \citet{phang2018sentence} explored supplementary training on intermediate tasks (STILT), \textit{i.e.}, training on a data-rich intermediate task before fine-tuning on the target task. STILT improves performance on the target task and stabilizes the fine-tuning process. \citet{pruksachatkun-etal-2020-intermediate} and \citet{vu-etal-2020-exploring} further investigated when and why intermediate task transfer works.
These studies mainly focus on transferability between specific \textit{source-target pairs}, while we consider a more general setting of transferring within and beyond a \textit{group} of NLP tasks. 



\section{Problem Setting}
Our goal is to better understand multi-task learning with mixture-of-experts models with an explicit routing mechanism.
We also hypothesize that such models help improve the model's capability to generalize/adapt to new tasks. 
Our problem setting closely resembles CrossFit \cite{ye-etal-2021-crossfit}.
In the following, we introduce data usage (\S\ref{ssec:data}), training procedure (\S\ref{ssec:training_precedure}), and evaluation protocol (\S\ref{ssec:evaluation_protocol}).

\subsection{Data Usage}\label{ssec:data}
Assume that we have a collection of diverse NLP tasks $\mathcal{T}$, partitioned into two non-overlapping sets $(\mathcal{T}_{train}, \mathcal{T}_{test})$. These sets are also referred to as (Meta-Train, Meta-Test). $\mathcal{T}_{train}$ is mainly used for multi-task learning; $\mathcal{T}_{test}$ is used to quantify the model's adaptability to new tasks.
Each task $T\in\mathcal{T}$ has three subsets, i.e., $T=(D_{train}, D_{dev}, D_{test})$. Additionally, we assume that all tasks are cast to a unified text-to-text format, \textit{i.e.}, $D =\{(x, y)\}$, where $x$ is the input text sequence, and $y$ is the output text sequence.


\subsection{Training Procedure}\label{ssec:training_precedure}
The training procedure has two stages: (1) an \textbf{upstream learning stage} for multi-task learning on $T_{train}$, to develop the skills that are needed to solve different tasks; and (2) a \textbf{downstream fine-tuning stage} on $T_{test}$, for evaluating the model's ability to adapt to new tasks.
During the upstream learning stage, the model is expected to be trained for multi-task learning with the $D_{train}$ from tasks in $\mathcal{T}_{train}$. $D_{dev}$ for tasks in $\mathcal{T}_{train}$ will be used for hyperparameter tuning and model selection. 
During the downstream fine-tuning stage, the model will be fine-tuned on each task in $\mathcal{T}_{test}$ respectively. $D_{train}$ will be used for fine-tuning, $D_{dev}$ for validation, and $D_{test}$ for reporting the final performance.

\subsection{Evaluation Protocol}\label{ssec:evaluation_protocol}
Each task in $\mathcal{T}$ has a pre-defined evaluation metric. For example, F1 score for classification tasks, and accuracy for multi-choice QA tasks.
During the upstream learning stage, for simplicity, the model is validated on the \textit{average} $D_{dev}$ performance on all tasks in $\mathcal{T}_{train}$, and we report \textit{average} $D_{dev}$ performance and $D_{test}$ performance. 
During the downstream fine-tuning stage, we compare the model's performance to the baseline of fine-tuning a vanilla transformer (without upstream learning), and compute the average relative performance gain (ARG) as our evaluation metric. More details about the baselines and ARG are deferred to \S\ref{sec:few-shot}.

\section{Task-level MoE Transformers}\label{ssec:model}

\begin{figure}
    \centering
    \includegraphics[width=0.5\textwidth]{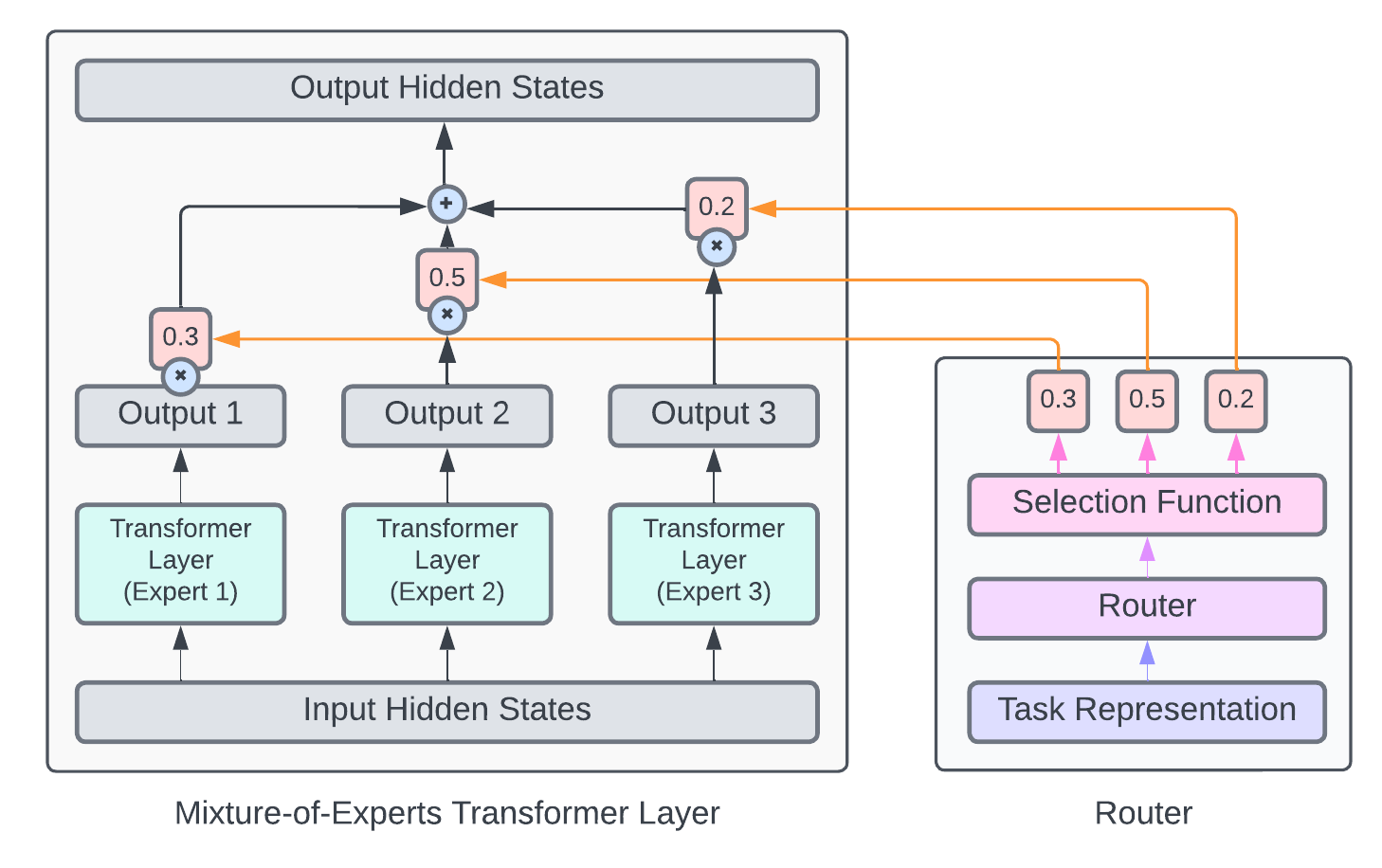}
    \caption{\textbf{Task-level Mixture-of-experts Transformer models used in this study.} \textbf{Right:} A router takes in a task representation and make decisions on expert selection. \textbf{Left:} the weighted sum of the outputs from each expert are considered the final output for this layer.}
    \label{fig:arch}
\end{figure}

Recall that our goal is to better elicit transferable skills during multi-task learning, and understand how those skills contribute the model performance. 
For this purpose we develop a mixture-of-experts variant of text-to-text transformer models, conditioning on task representations. 
The model contains two major components: (1) \textbf{a router} that selects and decides which experts to use for each task in each layer, based on its task representation; (2) \textbf{a collection of experts} that are dynamically composed into a final model based on the router selection. 
See Fig.~\ref{fig:arch} for a detailed illustration.


In the following, we introduce the router and the experts in more details. Note that we provide a general description in this section, and leave specific design choices in \S\ref{ssec:design_choices} for empirical comparison.

\paragraph{Collection of Experts.} In an \textit{original} implementation of text-to-text models \cite{t5,lewis-etal-2020-bart}, there are $n$ transformer layers stacked and executed sequentially. The first $n/2$ layers are encoder layers and the last $n/2$ layers are decoder layers.
In \textit{our variant} of transformer models, we copied each layer for $m$ times, resulting in $m*n$ experts in total. We refer to the $j$-th expert in the $i$-th layer as $E^{(i,j)}$. Note that we assume that each transformer block is an expert, which is different from \citet{Kudugunta21Beyond}. This is to make whole model dynamic and conpositional.

\paragraph{Router.} For a given task $T_{k}\in\mathcal{T}$, with $k$ as its task index, the router first takes the task representation ($\mathbf{T}_k$) from a look-up embedding table ($\mathbf{T}$). The router network outputs a matrix $\mathbf{L}\in \mathbb{R}^{m\times n}$, where $\mathbf{L}_{i,j}$ represents the logits of using expert $E^{(i,j)}$ in layer $i$. $\mathbf{L}$ goes through a selection function $f$ to normalize the routing decisions in each layer, resulting in a final decision matrix $\mathbf{D}\in \mathbb{R}^{m\times n}$. 

\paragraph{Task-level MoE Transformers.} We use the decision matrix $\mathbf{D}$ from the router to control the computation conducted by the experts. More specifically, in layer $i$, given input hidden states $\mathbf{h}^{(i)}_{in}$, the output $\mathbf{h}^{(i)}_{out}$ would be the weighted sum of all experts in the layer, and the weights are specified in $\mathbf{D}_{i,\cdot}$, \textit{i.e.},
\begin{equation}
    \mathbf{h}^{(i)}_{out}=\sum_{j=1}^{m} \mathbf{D}_{i,j} E^{(i,j)}(\mathbf{h}^{(i)}_{in})\label{eq:moe}
\end{equation}

\section{Applying Task-level MoE Models to Multi-task Learning}
\label{sec:multitask}

In our pilot studies, we found it is non-trivial to train these mixture-of-experts models properly and effectively. 
In this section, we present a detailed empirical study on baselines and design choices.
We first introduce experiment details in \S\ref{ssec:experiment_details}. 
We then start with investigating simple baselines such as random or average routing (\S\ref{ssec:baselines}), which will help navigate our experiments on \textit{learning} task-level MoE models. 
In \S\ref{ssec:design_choices} we introduce different variants we experiment with for learning task-level MoEs, and we summarize our findings in \S\ref{ssec:multitask_results}.

\subsection{Experiment Details}\label{ssec:experiment_details}
\paragraph{Data.} 
We previously discussed that a collection of diverse NLP tasks is required for the purpose of our study (\S\ref{ssec:data}). 
In our experiments, we use the task collection in CrossFit \cite{ye-etal-2021-crossfit}, which contains NLP tasks covering a wide range of task formats, goals and domains. We use its random task partition, with 120 tasks in $T_{train}$ and 18 tasks in $T_{test}$.
All tasks are converted to a unified text-to-text format and sub-sampled to be few-shot\footnote{For classification tasks, there are 16 examples per task in $D_{train}$; for non-classification tasks, $D_{train}$ has 32 examples.}.
Details about the tasks are listed in Appendix~\ref{app:all_tasks}-\ref{app:random_task_partition}.

\paragraph{Model and Its Initialization.}
We previously introduced the model architecture of task-level MoEs in \S\ref{ssec:model}. In our experiments, the model is instantiated with the pre-trained BART-Base model \cite{lewis-etal-2020-bart}, a 12-layer encoder-decoder transformer model ($n=12$). All $m$ experts in layer $i$ are initialized from the $i$-th layer of the BART-Base model. Additionally we add a Gaussian noise with variance of 1e-8 to the weights of each expert to avoid symmetry. We manually set the number of experts per layer $m=3$ to allow sufficient flexibility while maintain a tractable model size.
\paragraph{Training Details.} Deferred in Appendix \ref{app:multitask_details}.

\subsection{Investigation on Baselines}\label{ssec:baselines}
Before we experiment with learning routers, we first launch a series of baseline experiments related to the task-level MoE architecture. The goal is to get insights to help us better design our final model. We experiment with \colorbox{gray!10}{(1) Vanilla transformer}, where mixture-of-experts are not involved; \colorbox{gray!10}{(2) Instance-level random} routing, where the routes are randomly sampled for \textit{each instance} during the forward pass; \colorbox{gray!10}{(3) Task-level random} routing, where routes are sampled for \textit{each task} once before training; \colorbox{gray!10}{(4) Average} routing, where each experts were assigned the same weight in Eq.~(\ref{eq:moe}), \textit{i.e.}, $\mathbf{D}_{i,j}=1/3$. 
For (2) and (3), we try random selecting either one or two out of the three experts in each layer (denoted as ``1/3'' and ``2/3''). In the case of ``2/3'', the output is the average of the outputs produced by the activated experts.

\paragraph{Findings.} 
Performance of these baseline models are in top 2 sections in Table~\ref{tab:3runs}. We also plot the dev loss and performance curves during vanilla baseline training in Fig.~\ref{fig:vanilla_training} in Appendix \ref{app:discrepancy}. 
We have the following findings.

\textbf{(1)} In Fig.~\ref{fig:vanilla_training}, we found that dev losses dip in the early phase of training, then gradually rise. Meanwhile, the dev performance continue to increase. This is an important lesson learned for comparing different design choices: the simple and faster heuristic of model selection based on dev loss may be sub-optimal. We hypothesize this is because the text generation loss may not align well with the final evaluation metric\footnote{This finding is relevant to \citet{csordas-etal-2021-devil} which advocates proper validation protocol.}.

\textbf{(2)} All random routing methods (except for ``Random Task 2/3'') leads to worsened performance compared to vanilla transformer baselines. This suggests that introducing sparsity and routing mechanism into transformer models naively can in fact hurt performance. This may be due to underfitting (the number of examples routed to each expert is reduced) or asynchronism in optimization (a different collection of experts is activated and updated at each optimization step).

\textbf{(3)} The observation that Random Task Routing (2/3) is better than Vanilla and Average Routing suggests that task interference exists in multi-task models with \textit{fully} shared parameters, and allowing task-specific computations (as in Random Task 2/3) can be helpful.
The observation that Random Task 2/3 is better than 1/3 suggests that performance is highly sensitive to the portion of shared vs. task-specific parameters.
There is a fine line between MoE mechanism being helpful or being intrusive, adding difficulty to \textit{training} MoE models.

\subsection{Investigation on Design Choices}\label{ssec:design_choices}
In the following we describe the key design choices we compared in training task-level MoEs.

\paragraph{\colorbox{red!10}{Expert Selection.}} 
The selection function $f$ is responsible for normalizing and discretizing (if necessary) the logit output of router network into final decisions. We consider three variants: 
\colorbox{red!10}{(a) Softmax}, the default design in most MoE models. 
\colorbox{red!10}{(b) Gumbel-Softmax} \cite{jang2016categorical}, which add gumbel-distributed noise to the logits and promote discrete decisions. 
\colorbox{red!10}{(c) Gumbel-Softmax ST}, where ST stands for straight-through estimator. 
For (b) and (c), we apply the temperature annealing mechanism to encourage exploration in the beginning of training.

\paragraph{\colorbox{blue!10}{Router Architecture.}}
Router is a key component for our MoE model which computes the logits of selecting experts based on input task representations (see \S\ref{ssec:model}). We consider three router architecture with different complexities:
\colorbox{blue!10}{(d) MLP}, which contains two dense layers separated by GELU activation. 
\colorbox{blue!10}{(e) Bi-LSTM}, which takes the sum of the task representation and a positional embedding as input at each time step (\textit{i.e.}, layer). One linear layer is used to project the LSTM states to routing decisions.
\colorbox{blue!10}{(f) Transformer} \cite{vaswani2017attention}, which takes the same input as Bi-LSTM and applies one single transformer encoder layer.

\paragraph{\colorbox{green!10}{Task Representations.}}
\citet{vu-etal-2020-exploring} suggest that pre-computed task representations contain rich information for predicting task transferability. Here we consider incorporating these task representations as the initialization for the look-up embedding table $\mathbf{T}$ in our model (\S\ref{ssec:model}). In particular, we consider: 
\colorbox{green!10}{(g) Random}, which initialized every task representation with a randomly initialized 768d vector. 
\colorbox{green!10}{(h) TextEmb}, which is produced by encoding the input text with a pre-trained BART-Base model and taking the representations of the last encoder layer. We tried both the average representation of all tokens in the sequence (AVG) and BOS token representation.
\colorbox{green!10}{(i) FT-TextEmb}, which is mostly identical to (h), despite that the BART-Base model is first fine-tuned on the $D_{train}$ of the current task. 
\colorbox{green!10}{(j) Fisher-TaskEmb} \cite{vu-etal-2020-exploring}, which is the diagonal of fisher information of the trainable parameters in a model. We use adapter \cite{Houlsby2019ParameterEfficientTL} fine-tuning on $D_{train}$ and compute the fisher information on these adapter parameters to avoid expensive computations.

\paragraph{\colorbox{teal!10}{Freezing Task Representations.}}
Since adaptability to unseen task will be considered in later parts of this study, we further consider between \colorbox{teal!10}{(k) not freezing} and \colorbox{teal!10}{(l) freezing} the task representations during multi-task learning. We conjecture that the structure of seen task representations may be changed after multi-task learning, while the unseen task representations may not reflect the change; hence the freezing variant.

\begin{table}[t]
\centering
\scalebox{0.65}{
\begin{tabular}{lccc}
\toprule
\textbf{Model}     & \textbf{Compute} & \textbf{Dev} (\%) & \textbf{Test} (\%) \\
\midrule
\rowcolor{gray!30}
\multicolumn{4}{l}{\textbf{Vanilla Transformers}}\\
\midrule
(1) BART-Base  & 1x & 54.47$\pm$0.05& 48.93$\pm$0.23 \\
(1) BART-Large & - & 58.10$\pm$0.20 & 54.06$\pm$0.22 \\
\midrule
\rowcolor{gray!30}
\multicolumn{4}{l}{\textbf{Baselines}} \\\midrule
(2) Random Inst. Routing (1/3) & 1x & 47.50$\pm$0.20 & 41.87$\pm$0.76 \\
(2) Random Inst. Routing (2/3) & 2x & 44.81$\pm$1.76 & 38.48$\pm$1.00 \\
(3) Random Task Routing (1/3) & 1x & 52.89$\pm$0.57 & 47.27$\pm$0.35 \\
(3) Random Task Routing (2/3) & 2x & \textbf{55.35$\pm$0.23} & \textbf{50.44$\pm$0.29} \\
(4) Average Routing (3/3) & 3x & 54.61$\pm$0.11 & 50.02$\pm$0.19 \\
\midrule
\rowcolor{gray!30}
\multicolumn{4}{l}{\textbf{Task-level Mixture-of-Experts}} \\\midrule
\colorbox{red!10}{(c)}+\colorbox{blue!10}{(d)}+\colorbox{green!10}{(g)}+\colorbox{teal!10}{(k)}+\colorbox{brown!10}{(n)} & 1x & \textbf{55.28$\pm$0.12} & \textbf{50.52$\pm$0.38} \\
\colorbox{red!10}{(c)}+\colorbox{blue!10}{(d)}+\colorbox{green!10}{(j)}+\colorbox{teal!10}{(k)}+\colorbox{brown!10}{(m)} & 1x & 53.07$\pm$0.45 & 48.16$\pm$0.34 \\
\colorbox{red!10}{(c)}+\colorbox{blue!10}{(d)}+\colorbox{green!10}{(j)}+\colorbox{teal!10}{(l)}+\colorbox{brown!10}{(m)} & 1x & 53.06$\pm$0.19 & 47.64$\pm$0.79 \\
\colorbox{red!10}{(c)}+\colorbox{blue!10}{(d)}+\colorbox{green!10}{(j)}+\colorbox{teal!10}{(l)}+\colorbox{brown!10}{(n)} & 1x & \textbf{55.40$\pm$0.08} & \textbf{50.39$\pm$0.68} \\
\bottomrule
\end{tabular}
}
\caption{\textbf{Performance on baselines and selected models.} Average performance on $D_{dev}$/$D_{test}$ over tasks in $\mathcal{T}_{train}$ are reported. Average and standard dev. are computed based on runs with three different random seeds.}\label{tab:3runs}
\end{table}

\begin{table}[t]
\centering
\vspace{-0.2cm}
\scalebox{0.65}{
\begin{tabular}{lc|lc}
\toprule
\textbf{Model}     & \textbf{Dev} (\%) & \textbf{Model}     & \textbf{Dev} (\%)\\
\midrule
\cellcolor{red!10} 
Expert Selection & \cellcolor{red!10} & \cellcolor{green!10} Task Repr. (cont.)& \cellcolor{green!10}\\
(a) Softmax & 40.93 & (i) FT-TextEmb-BOS & 52.93 \\
(b) Gumbel-Softmax & 52.02 & (i) FT-TextEmb-AVG & 53.29 \\
(c) Gumbel-Softmax ST & 53.14 & (j) Fisher-TaskEmb & 53.51 \\
\midrule
\cellcolor{blue!10}
Router Architecture & \cellcolor{blue!10} & \cellcolor{teal!10} Freeze Task Repr. & \cellcolor{teal!10}\\
(d) MLP & 53.14 & (k) Not Freezing & 53.51 \\
(e) LSTM & 53.55 & (l) Freezing & 53.37\\
(f) Transformer & 53.13 & - & -\\
\midrule
\cellcolor{green!10}
Task Repr. & \cellcolor{green!10} & \cellcolor{brown!10} Two-stage Training & \cellcolor{brown!10}\\
(g) Random & 53.14 & (m) Use one stage & 53.51\\
(h) TextEmb-BOS & 52.51 & (n) Use two stages & 55.36\\
(h) TextEmb-AVG & 53.30 & - & -\\
\bottomrule
\end{tabular}
}
\caption{\textbf{Investigation on Design Choices.} By default the model uses \colorbox{red!10}{(c)}+\colorbox{blue!10}{(d)}+\colorbox{green!10}{(g)}+\colorbox{teal!10}{(k)}+\colorbox{brown!10}{(m)} when comparing different choices in each colored section.}
\vspace{-0.2cm}
\label{tab:design_choices}
\end{table}

\paragraph{\colorbox{brown!10}{Two-stage Training.}}
In \S\ref{ssec:baselines}, we find that introducing routing mechanism naively may lead to worsened performance. Also, average routing is stable and achieves competitive performance. Based on these observations, we design a two-stage training strategy to combine the benefits of both methods. In the first stage, the model jointly learns the router and the experts. In the second stage, the experts are re-initialized from BART's pre-trained weights, and the routes gradually transforms from average routing to the learned routes by controlling the temperature used in the softmax function. 
As a result, in the beginning of the training, the temperature is set to be high, so the router is functioning like average routing; during the training process, the temperature decreases gradually, and the router will give more discrete routing decisions.


\subsection{Results and Findings}
\label{ssec:multitask_results}
We first present the performance of variants mentioned above in Table~\ref{tab:design_choices}. For the best-performing model variants, we run three times with different random seeds to reduce variance in performance (Table~\ref{tab:3runs}, Bottom). We have the following observations. 
\textbf{(1) What helps?} We found that the choice of \colorbox{red!10}{selection function} and the \colorbox{brown!10}{two-stage learning} procedure are important for training task-level MoEs. Gumbel-Softmax with straight-through estimator achieves the best performance among the three choices.\footnote{See Appendix~\ref{app:softmax} for further investigation.} Two-stage training helps improve performance by 1.8\%.\footnote{We also use heterogeneous batching \cite{aghajanyan-etal-2021-muppet} and two-speed learning rate \cite{ponti2022combining} in our model as recommended by these works.}
\textbf{(2) What doesn't help?} We did not observe significant difference with choices in \colorbox{blue!10}{router architecture} or \colorbox{green!10}{task representation initialization}.
Supposedly, LSTMs and transformers are able to capture relations more complicated than MLPs, and pre-computed task representations carry richer information about the task than random initialization.
This unexpected observation suggests that the router struggle to leverage task-level information with the current training methods and supervision signals.
\textbf{(3) Comparing with the baselines.} Our best task-level MoE using random initialized task representations (\colorbox{red!10}{(c)}+\colorbox{blue!10}{(d)}+\colorbox{green!10}{(g)}+\colorbox{teal!10}{(k)}+\colorbox{brown!10}{(n)}) can rival the best baselines in \S\ref{ssec:baselines} (Random Task Routing 2/3), while using half of its computation in a forward pass. With careful design, task-level MoEs are beneficial for multi-task learning.

\begin{figure*}[!t]
\vspace{-0.5cm}
    \centering
    \includegraphics[width=0.92\textwidth]{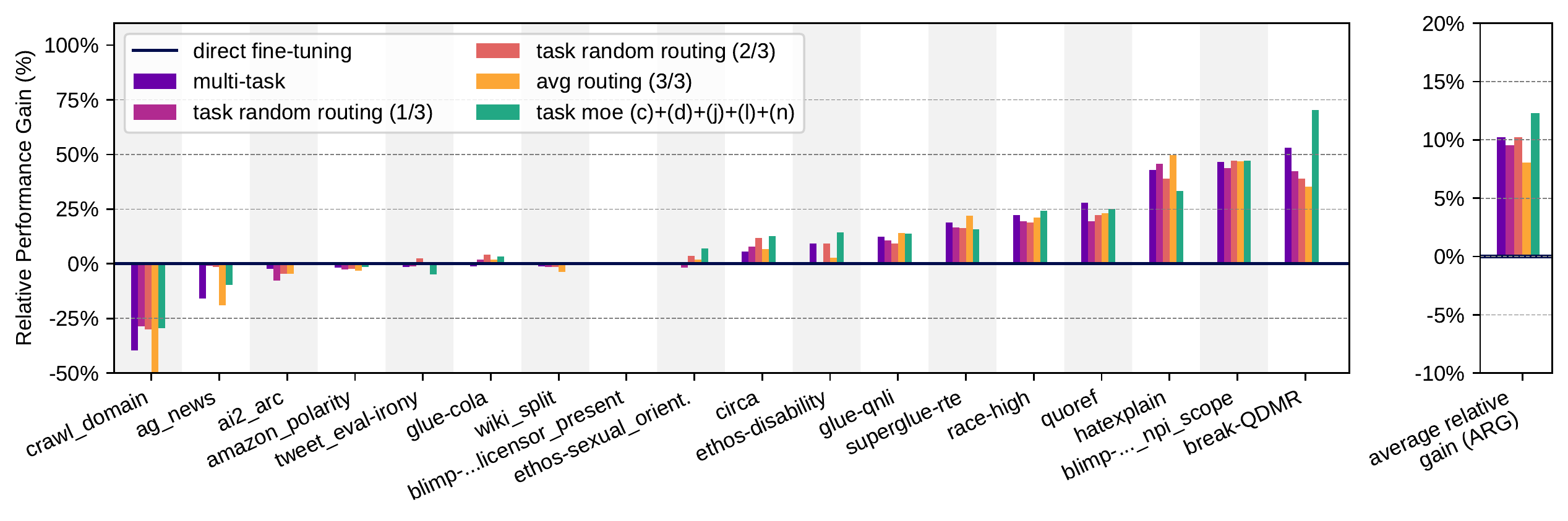}
    \vspace{-0.2cm}
    \caption{\textbf{Few-shot Performance on Unseen Tasks.} Bar heights represent relative performance gain over directly fine-tuning a pre-trained BART-Base model. The right-most bars are the average performance gain.}
    \label{fig:few-shot}
\end{figure*}

\begin{table*}[!t]
\vspace{-0.2em}
\hspace{-0.8em}
\centering
\scalebox{0.7}{
\begin{tabular}{r||c|c|c|c|c|c|c|c|c|c||c|c}
\toprule
Main models & anli\_r3 & HellaSwag  &  cb         &    wic          & wsc    & winogrande    & arc-chan. & obqa  & piqa & SQuADv2 & AVG & ARG\\
\midrule
Multi-task BART-Base         & 27.6 & 22.0 & 44.6 & 43.1 & 57.5 & 52.7 & 23.1 & 26.2  & 26.4  & 14.6 & 33.7 & -   \\
Random Task Routing (2/3) & 23.7 & 14.5 & 19.3 & 37.8 & 45.2 & 49.0 & 14.5 &  18.5 & 3.5 & 9.1 & 23.5 & -33.6\\
\colorbox{red!10}{(c)}+\colorbox{blue!10}{(d)}+\colorbox{green!10}{(h)}+\colorbox{teal!10}{(l)}+\colorbox{brown!10}{(m)}        &   33.7 & 20.7 & 43.6 & 40.4 & 50.2 & 46.8 & 11.8 & 18.1 & 22.4 &18.6 & 32.2 &  -8.3        \\
\colorbox{red!10}{(c)}+\colorbox{blue!10}{(d)}+\colorbox{green!10}{(h)}+\colorbox{teal!10}{(l)}+\colorbox{brown!10}{(n)}               & 32.0 & 23.7 & 44.3 & 43.4 & 56.5 & 52.2 & 21.1 & 28.5  &30.2 & 17.6 &34.9 & 5.6\\
\bottomrule

\end{tabular}
}
\caption{\textbf{Zero-shot Performance on Unseen Tasks}. Accuracy (\%) on the test set of 10 unseen tasks. We compare the AVG and calculate the ARG of routing model \colorbox{red!10}{(c)}+\colorbox{blue!10}{(d)}+\colorbox{green!10}{(h)}+\colorbox{teal!10}{(l)}+\colorbox{brown!10}{(m)} and \colorbox{red!10}{(c)}+\colorbox{blue!10}{(d)}+\colorbox{green!10}{(h)}+\colorbox{teal!10}{(l)}+\colorbox{brown!10}{(n)} over multi-task BART-Base. The former routing model uses one-stage training while the latter uses two-stage straining.}
\vspace{-0.2cm}
\label{tab:zeroshot_each}
\end{table*}

\section{Generalizing to Unseen Tasks}
\label{sec:few-shot}
We hypothesize that task-level MoE models can recombine the learned skills effectively when they encounter new tasks. In \S\ref{ssec:few-shot} we evaluate the models obtained in \S\ref{sec:multitask} on adapting to new tasks in a few-shot learning setting. In \S\ref{ssec:zero-shot} we further extend our method to a zero-shot learning setting and test it on the P3 dataset \cite{sanh2022multitask}.

\subsection{Few-shot Adaptation}\label{ssec:few-shot}

\paragraph{Compared Methods.} We use the following models as initialization for few-shot fine-tuning on unseen tasks ($T_{test}$).
\textbf{(1) Direct Fine-tuning.} For each unseen task, we fine-tune the off-the-shelf BART-Base model with its $D_{train}$.
\textbf{(2) Multi-task BART.} We take the multi-task BART-Base from \S\ref{sec:multitask} as initialization and fine-tune the model on $D_{train}$.
\textbf{(3) Baseline Routing BART.} We re-use the models using random task routing (1/3, 2/3) and average routing in \S\ref{sec:multitask}.
\textbf{(4) Learned Routing BART.} We take the
\colorbox{red!10}{(c)}+\colorbox{blue!10}{(d)}+\colorbox{green!10}{(j)} +\colorbox{teal!10}{(l)}+\colorbox{brown!10}{(n)} model
from \S\ref{sec:multitask}. 
This models uses fisher information as the task representation \colorbox{green!10}{(j)} and the representations for seen tasks are frozen \colorbox{teal!10}{(l)} during multi-task learning. For the unseen task, we first compute its fisher information based on $D_{train}$ and feed it to the learned router to select experts. We then fine-tune the selected experts on $D_{train}$.

\paragraph{Data and Evaluation.}
We use the 18 unseen tasks specified in CrossFit random partition in \citet{ye-etal-2021-crossfit}\footnote{We exclude Free-base QA and Yelp Polarity from the evaluation as performance is unusually unstable on these tasks.}.
We first obtain the performance of fine-tuning the pre-trained BART-Base model as the baseline. 
Then we compute and report the average relative gain (ARG) over pre-trained BART for the multi-task BART and routing BART methods. For example, if fine-tuning pre-trained BART achieves 50\% accuracy on task A and 80\% F1 on task B, and fine-tuning multi-task BART achieves 80\% accuracy on task A and 60\% F1 on task B, the ARG would be the average of $(80\%-50\%)/50\%$ and $(60\%-80\%)/80\%$, which equals to $17.5\%$.

\paragraph{Results.}
We present the performance gains on individual tasks and their average in Fig.~\ref{fig:few-shot}. 
Multi-task BART remains a strong baseline, achieving an ARG of 9.74\%. 
Random task routing (2/3) and average routing baselines achieves 10.21\% and 8.06\% respectively.
Our task-level MoE model \colorbox{red!10}{(c)}+\colorbox{blue!10}{(d)} +\colorbox{green!10}{(j)}+\colorbox{teal!10}{(l)}+\colorbox{brown!10}{(n)} achieves the best average performance gain (12.30\%), which is 2.6\% higher than the multi-task BART. We observe that negative transfers are alleviated and few-shot performance are improved compared to the baselines for many tasks. This suggest that our task-level MoE model is learning reusable experts and meaningful routes.

\subsection{Zero-shot Generalization}\label{ssec:zero-shot}
In this section, we modify our proposed method to zero-shot learning settings where each unseen task has no labeled data. We use Public Pool of Prompts (P3) dataset as our testbed \citep{sanh2022multitask}.

\paragraph{Data.} Following \citet{sanh2022multitask, bach-etal-2022-promptsource}, we use the prompt templates to change texts from various NLP tasks into a unified text-to-text formats. Specifically, we have 36 upstream tasks for $\mathcal{T}_{train}$, and 10 tasks for $\mathcal{T}_{test}$. We use accuracy as the evaluation metric. We report both the average performance on $\mathcal{T}_{test}$ (AVG), and the average performance gain (ARG) described in \S\ref{ssec:few-shot}.

\paragraph{Compared Methods.} For all the models, we train on the $D_{train}$ for all tasks in $\mathcal{T}_{train}$, and directly test the model on $D_{test}$ for each task in $\mathcal{T}_{test}$. 
We mainly compare four methods: \textbf{(1)} Multi-task BART-Base. \textbf{(2)} Random Task Routing (2/3). \textbf{(3)} We train a new \colorbox{red!10}{(c)}+\colorbox{blue!10}{(d)}+\colorbox{green!10}{(h)}+\colorbox{teal!10}{(l)}+\colorbox{brown!10}{(m)} model on P3 data. \textbf{(4)} Similar to (3), we train a model with the configuration of \colorbox{red!10}{(c)}+\colorbox{blue!10}{(d)}+\colorbox{green!10}{(h)}+\colorbox{teal!10}{(l)}+\colorbox{brown!10}{(n)}. 
Note that in the zero-shot setting, we cannot use pre-computed task representations for unseen tasks based on labeled examples (as described in \S\ref{ssec:design_choices}). Therefore for \colorbox{green!10}{(h) TextEmb} used in \textbf{(3)} and \textbf{(4)}, we encode prompt templates as the auxiliary task information. More details are in Appendix~\ref{app:zeroshot_setting}.

\paragraph{Results.} 
We present the results in Table~\ref{tab:zeroshot_each}. 
Our findings are:
\textbf{(1)} Compared to the multi-task BART-base baseline with an AVG of 33.7\%, our routing model (4) achieves a higher AVG (34.9\%) and a positive ARG (5.6\%).
This demonstrates the model's improved generalization ability to novel tasks in the zero-shot setting. 
\textbf{(2)} The gap between model (3) and model (4) shows that the two-stage training strategy is essential in the zero-shot setting as well.
\textbf{(3)} Different from the findings in the few-shot setting, Random Task Routing (2/3) has a negative ARG (-33.6\%). Without labeled data in unseen tasks, random routing cannot actively select relevant experts or update model parameters, resulting in worsened performance. In contrast, task-level MoE has the flexibility to select relevant experts and achieves better performance.

\section{Interpreting the Routes and Experts}
\label{sec:interpret}

\subsection{Learning Dynamics of the Routes}
\begin{figure}[!t]
    \centering
    \includegraphics[width=0.49\textwidth, height=0.42\textwidth]{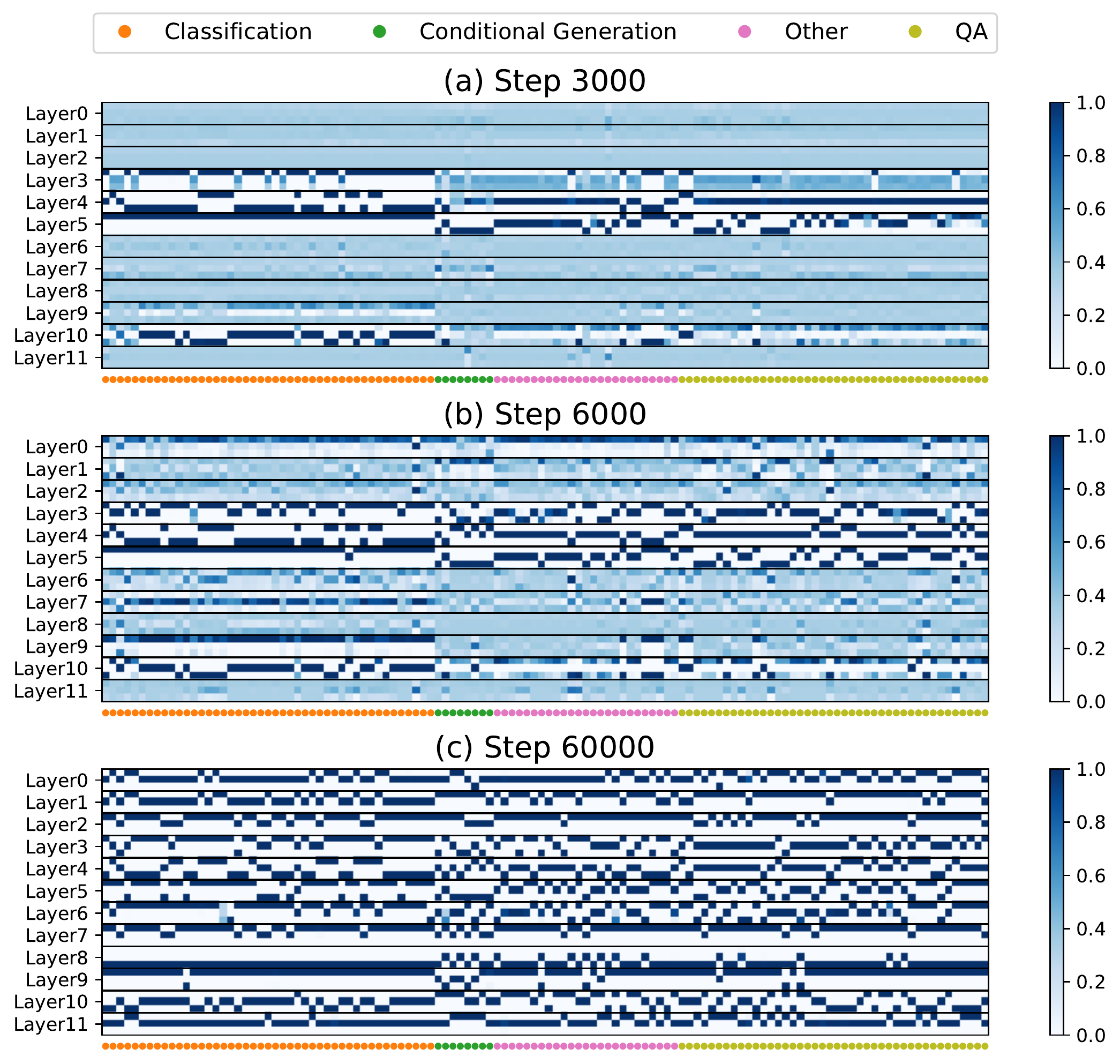}
    \caption{\textbf{Routing Decisions Learned During Multi-task Learning} (\colorbox{red!10}{(c)}+\colorbox{blue!10}{(d)}+\colorbox{green!10}{(g)}+\colorbox{teal!10}{(k)}+\colorbox{brown!10}{(m)}). The router is able to distinguish classification tasks from other types of tasks after 3000 steps of the training. It then gradually learns more fine-grained patterns.}
    \label{fig:dynamics}
\end{figure}

We visualized the learned routing decisions of the \colorbox{red!10}{(c)}+\colorbox{blue!10}{(d)}+\colorbox{green!10}{(g)}+\colorbox{teal!10}{(k)}+\colorbox{brown!10}{(m)} model trained on CrossFit data in Fig.~\ref{fig:dynamics}. Note that \colorbox{green!10}{(g)} represents that the task representations are randomly initialized and learned spontaneously during multi-task learning. We observe that distinct patterns for classification and generation tasks emerge in the early stage of the training (step 3000). These patterns transition from coarse-grained to fine-grained gradually in the training process. These observations align with our expectation that task-level MoEs are learning to share parameters for similar tasks and avoid interference among dissimilar tasks.

\subsection{Correlation with Task Features}
To better understand the learned routing decisions, we investigate the relation between the routing decisions and manually-defined task features. In the following, we first describe the methodology of computing correlation, then describe the features we investigate, and finally describe our findings. 

\paragraph{Method.} 
For each task in $\mathcal{T}_{train}$, we first compute the routing decisions $\mathbf{D}\in \mathbb{R}^{m\times n}$ using the learned model. 
For each expert $E^{(i,j)}$, we consider the routing decision $\mathbf{D}_{i,j}$ of all tasks as a feature. 
Altogether, we have $m\times n$ features of dimension $|\mathcal{T}_{train}|$ (the number of tasks). 
Additionally, we have $t$ manually-defined features on all tasks, giving $t$ features of dimension $|\mathcal{T}_{train}|$. 
We compute Pearson correlation coefficient between each pair of learned routing decisions and manual feature, resulting in a $\mathbb{R}^{mn\times t}$ matrix quantifying the correlation between $m\times n$ experts and $t$ manual features.

\begin{table}[t]
\hspace{-0.2cm}
\scalebox{0.56}{
\begin{tabular}{lp{2.5cm}p{7cm}}
\toprule
\textbf{Feature Name} & \textbf{Example} & \textbf{Description}                                                                          \\\midrule
\rowcolor{gray!30}\multicolumn{3}{l}{Task Format}                                                                                                          \\
Extractive            & SQuAD, Race  & Output is always a substring of the input                                                     \\
Sentence Completion   &  HellaSwag, LAMA-Probes  & Requires the model to fill in a blank in the input or continue to generate based on the input \\\midrule
\rowcolor{gray!30}\multicolumn{3}{l}{Required Skills and Knowledge}                       \\
Linguistic            &  Blimp,~CoLA  & Tasks focusing on grammatical correctness, semantic equivalence and linguistic phenomenon     \\
Commonsense           &  CommonsenseQA  & Tasks testing commonsense knowledge and reasoning capabilities                    \\
Co-reference          &  Wino\_grande                & Tasks requiring co-reference resolution                                                       \\
Multi-hop Reasoning & DROP & Tasks requiring multi-hop/multi-step reasoning\\
Implicit Knowledge    & TriviaQA  & Tasks requiring world knowledge (acquired during pre-training)                                \\
Synthesize            & Break, XSum & Combining ideas and allowing an evolving understanding of text    \\\bottomrule                           
\end{tabular}
}
\caption{Additional Features on Format, High-level Skills and Knowledge.}\label{tab:hand-features}
\end{table}

\paragraph{Manual Features.} We consider the following features in our correlation study\footnote{We admit that several categorization criteria are subjective and they are by no means exhaustive for fully describing a task. We use these features mainly to quantify the relation between human understanding of tasks and the learned routes.}. The final feature table ($t \times |\mathcal{T}_{train}|$) is in Table ~\ref{tab:hand_features}.
\begin{itemize}[leftmargin=*, nosep]
\itemsep0em 
\item \textbf{Task Format.} We use the task categories provided in \citet{ye-etal-2021-crossfit}. The top-level labels include \ul{\texttt{Classification}}, \ul{\texttt{Question Answering}}, \ul{\texttt{Conditional Generation}}, and \ul{\texttt{Others}}. Tasks in each category are divided into sub-categories. For example, QA tasks are further categorized into \ul{\texttt{machine reading comprehension (MRC)}}, \ul{\texttt{multiple-choice QA}}, \ul{\texttt{closed-book QA}}, etc.

\item \textbf{Input/Output Length.} 
We classify tasks with into three features based on their average input length: \ul{\texttt{hasShortInput}} (shortest 25\%), \ul{\texttt{hasLongInput}} (longest 25\%), \ul{\texttt{hasMediumInput}} (remainder). We also classify tasks into three features based on their average output length: \ul{\texttt{hasShortOutput}} ($<$ 3 tokens), \ul{\texttt{hasLongOutput}} ($>$ 10 tokens), and \ul{\texttt{hasMediumOutput}} (remainder).

\begin{figure}[!t]
    \centering
    \hspace{-0.01cm}
    \includegraphics[width=0.48\textwidth]{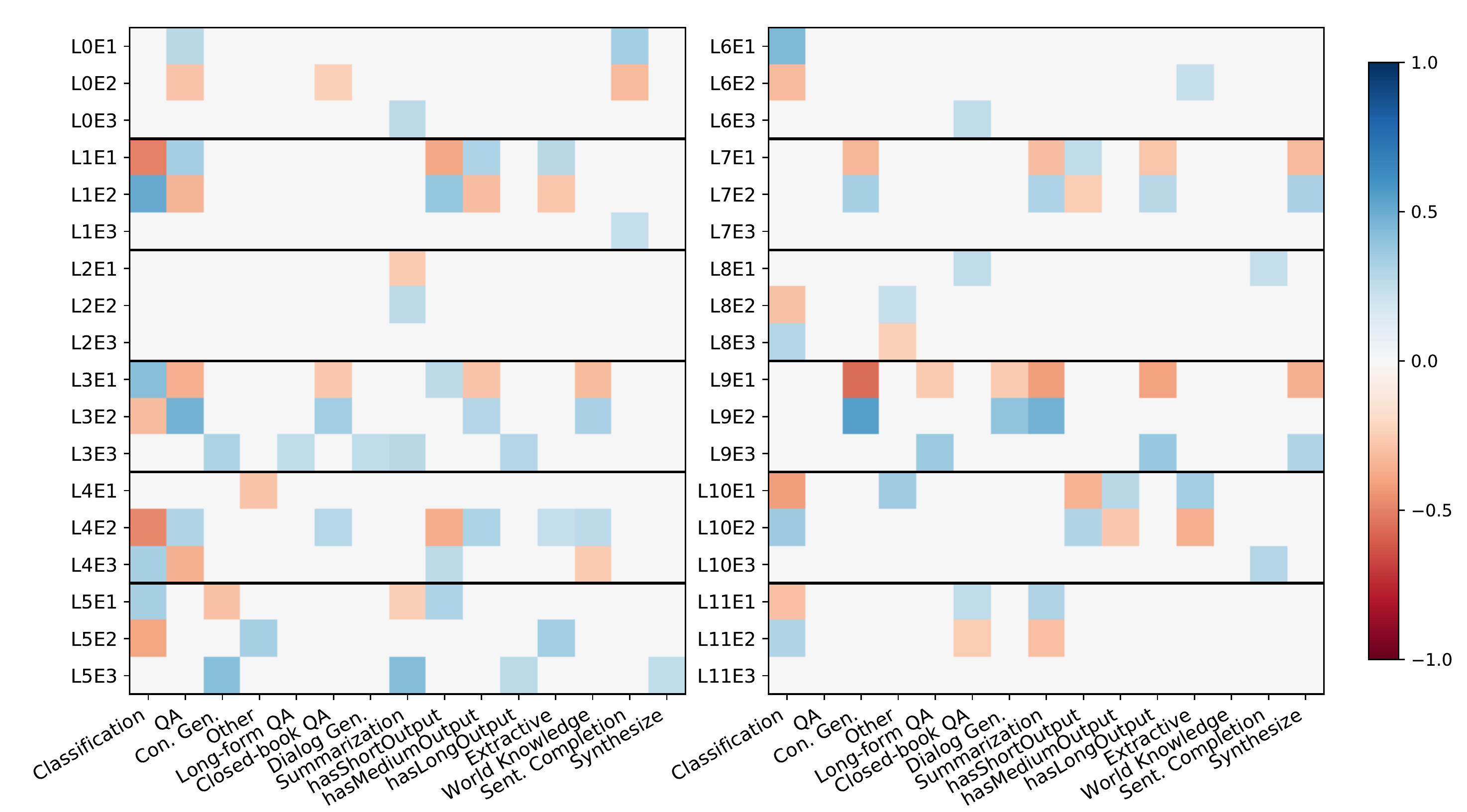}
    \caption{\textbf{Pearson Correlation Between Learned Routes and Selected Manual Features.} Correlation with $p$<$0.01$ are visualized. ``L0E1'' stands for expert 1 in layer 0. The correlation is computed based on a \colorbox{red!10}{(c)}+\colorbox{blue!10}{(d)}+\colorbox{green!10}{(g)}+\colorbox{teal!10}{(k)} model, where \colorbox{green!10}{(g)} means the task embedding table $\mathbf{T}$ is randomly initialized. This suggests that without prior knowledge of the tasks, the router can partially rediscover human categorization of tasks during multi-task learning. 
    }
    \label{fig:corr}
\end{figure}

\item \textbf{Text Domain.} 
We categorize tasks with into domains such as \ul{\texttt{Science \& Technology}}, \ul{\texttt{Social Network}}, \ul{\texttt{News}}, \ul{\texttt{Web}}, \ul{\texttt{Bio-Medical}}, \ul{\texttt{Review}}, \ul{\texttt{Dialog}}, and \ul{\texttt{Books}}.

\item \textbf{Granularity.} We categorize tasks into \ul{\texttt{Span-level}} (\textit{e.g.}, acronym identification); \ul{\texttt{Sentence-level}} (\textit{e.g.}, tweet classification); \ul{\texttt{Paragraph-level}} (\textit{e.g.}, news summarization) based on their main focus. This is different from input length.

\item \textbf{Additional Features: Format, High-level Skills and Knowledge\footnote{These features are mostly inspired by dataset papers such as SQuAD \cite{rajpurkar-etal-2016-squad}, BLiMP \cite{warstadt2019blimp}, MNLI \cite{williams-etal-2018-broad}, HotpotQA \cite{yang-etal-2018-hotpotqa}, CommonsenseQA \cite{talmor-etal-2019-commonsenseqa}.}.} We additionally describe several common task characteristics in Table~\ref{tab:hand-features}. These include whether a task is \ul{\texttt{Extractive}}, requires \ul{\texttt{Sentence Completion}}, or requires high-level skills such as \ul{\texttt{Co-reference}}.

\end{itemize}

\paragraph{Findings.} Results on selected features are visualized in Fig.~\ref{fig:corr}. Visualization of the complete pairs of expert and feature are in Fig.~\ref{fig:corr-random-full}-\ref{fig:corr-random-fisher}. We have the following observations: \textbf{(1)} There exists strong correlation between several pairs of routing decisions and manual features. For example, L1E2, L3E1, L6E1 are positively correlated with the feature of Classification, suggesting that these experts are likely to be selected for classification tasks. \textbf{(2)} The correlations are strongest with the top-level task category features (\textit{i.e.}, \ul{\texttt{Classification}}, \ul{\texttt{QA}}, \ul{\texttt{Conditional Generation}}), suggesting that the router may understand and categorize tasks in a way similar to us. \textbf{(3)} However, correlation does not imply causal relationships. The correlation patterns of \ul{\texttt{Classification}} and \ul{\texttt{hasShortOutput}} are similar, the same applies to \ul{\texttt{Conditional Generation}} and \ul{\texttt{hasLongOutput}}. We cannot conclude whether the router is making router decisions depending on output length, task format, or other hidden aspects.

\begin{table}[t]
\centering
\scalebox{0.65}{
\begin{tabular}{c|c|c|ccc}
\toprule
\textbf{Manual Feature}   & \textbf{Top3 Exp} & \textbf{Task} & \textbf{All} & \st{\textbf{Top1}} & \st{\textbf{Top3}} \\
\midrule
\multirow{3}{*}{Classification} &\multirow{3}{*}{\begin{tabular}[c]{@{}c@{}}L1E2\\ L6E1\\ L3E1\end{tabular}}& imdb  &   92.49   & 91.87   & 88.70   \\
                    && sms spam &   63.54   &  63.54 & 62.88   \\
                    && emo  &    82.06   &  65.46 & 16.22  \\
                    \midrule
\multirow{3}{*}{\begin{tabular}[c]{@{}c@{}}Conditional\\ Generation\end{tabular}} &\multirow{3}{*}{\begin{tabular}[c]{@{}c@{}}L9E2\\ L5E3\\ L7E2\end{tabular}}& gigaword & 30.00 & 26.51 & 17.91   \\
                    && aeslc  &  14.52  & 15.31 & 14.76   \\
                    && kilt\_wow  &  6.39  & 6.01 & 4.73   \\\midrule
\multirow{3}{*}{\begin{tabular}[c]{@{}c@{}}Closed-book\\ QA\end{tabular}} &\multirow{3}{*}{\begin{tabular}[c]{@{}c@{}}L3E2\\ L4E2\\ L6E3\end{tabular}}& kilt\_trex & 31.85 & 25.63 & 28.13   \\
                    && kilt\_zsre  &  13.13  & 11.25 & 9.38   \\
                    && numer\_sense &  34.38  & 33.75 & 20.00   \\
\bottomrule
\end{tabular}
}
\caption{\textbf{Performance when top correlated experts are disabled.} ``\st{Top1}'' means the most positively correlated expert is disabled. Performance gradually drops as more experts are disabled.}\label{tab:disable}
\end{table}

\begin{table}[t]
\centering
\scalebox{0.65}{
\begin{tabular}{c|c|cc|cc|cc}
\toprule
\textbf{Task}     & \textbf{All} & \st{\textbf{Top1}} & \st{\textbf{Top3}} & \st{\textbf{Rand1}} & \st{\textbf{Rand3}} & \st{\textbf{Least1}} & \st{\textbf{Least3}} \\\midrule
imdb     & 92.49 & 91.87 & 88.70 & 92.49 & 91.66 & 92.49 & 92.49 \\
sms spam & 63.54 & 63.54 & 62.88 & 63.54 & 63.53 & 63.54 & 63.54 \\
emo      & 82.06 & 65.46 & 16.22  & 82.06 & 64.13 & 82.06 & 82.06 \\\bottomrule
\end{tabular}
}
\caption{\textbf{Disabling top/least correlated experts and random experts.} The experts that positively correlate (Top1/Top3) with the ``classification'' feature contribute more to the performance than randomly selected or least correlated experts (Least1/Least3).}\label{tab:disable2}
\end{table}

\subsection{Expert Disabling Experiments}
We further examine the learned task-level MoE models by disabling experts during evaluation. By ``disabling'', we simply set the pre-softmax logit to be $-\infty$, so that the second-best expert in that layer will be selected instead. We hypothesize that if an expert corresponds to a critical skill required by a certain type of tasks, then disabling it should bring significant performance drop. 
\textbf{(1)} We select three manual features: \ul{\texttt{Classification}}, \ul{\texttt{Conditional Generation}}, \ul{\texttt{Closed-book QA}}, and select three tasks that belong to these categories. We select the top 3 experts that positively correlate with these features, and disable them during evaluation. Results are listed in Table~\ref{tab:disable}. As expected, these correlated experts are indispensable for the task performance. Performance gradually drops as more experts are disabled (All $\rightarrow$ \st{Top1} $\rightarrow$ \st{Top3}).
\textbf{(2)} For the three classification tasks we select, we further compare the performance when disabling most/least correlated experts and random experts. Results are presented in in Table~\ref{tab:disable2}. Results suggest experts that are positively correlated with the classification feature are more important to the final performance.
\textbf{(3)} We further take two classification tasks ($\diamondsuit$) and two closed-book QA tasks ($\heartsuit$), and consider disabling experts correlated with classification and closed-book feature. Results are shown in Table~\ref{tab:disable3}. Performance are not influenced significantly when experts relevant to other features are disabled.
\textbf{To conclude}, this set of experiments suggests that experts that positively correlate with a specific type of tasks are irreplaceable; they greatly contribute to the performance of that type of tasks.

\begin{table}[t]
\centering
\scalebox{0.65}{
\begin{tabular}{c|c|cc|cc}
\toprule
\textbf{Task}  & \textbf{All} & {$\diamondsuit$} \st{\textbf{Top1}} & {$\diamondsuit$} \st{\textbf{Top3}} & {$\heartsuit$} \st{\textbf{Top1}} & {$\heartsuit$} \st{\textbf{Top3}} \\\midrule
{$\diamondsuit$} imdb         & 92.49 & \cellcolor{green!10}91.87 & \cellcolor{green!10}88.70 & \cellcolor{red!10}92.49 & \cellcolor{red!10}92.49 \\
{$\diamondsuit$} emo          & 82.06 & \cellcolor{green!10}65.46 & \cellcolor{green!10}16.22 & \cellcolor{red!10}82.06 & \cellcolor{red!10}82.06 \\\midrule
{$\heartsuit$} kilt\_zsre   & 13.13 & \cellcolor{red!10}13.13 & \cellcolor{red!10}12.50 & \cellcolor{green!10}11.25 & \cellcolor{green!10}9.38 \\
{$\heartsuit$} numer\_sense & 34.38 & \cellcolor{red!10}34.38 & \cellcolor{red!10}34.38 & \cellcolor{green!10}33.75 & \cellcolor{green!10}20.00 \\\bottomrule
\end{tabular}
}
\caption{\textbf{Disabling experts associated with different task categories.} {$\diamondsuit$}=Classification, {$\heartsuit$}=Closed-book QA. Performance does not drop significantly when experts relevant to other features are disabled (red area).}\label{tab:disable3}
\end{table}

\section{Conclusions}
Inspired by how humans accumulate skills from past experience and re-use them to solve new tasks, in this paper,
we develop and conduct extensive experiments with transformer-based task-level mixture-of-expert (MoE) models, in hope to provide new insights on multi-task learning and cross-task generalization in NLP. 
Firstly, we empirically investigate importance design choices and quantify their influence on final model. 
Secondly, in both few-shot and zero-shot settings, we demonstrate that task-level mixture-of-expert models are better at generalizing to new tasks. 
Finally, by conducting a detailed analysis on the routing decisions, we find they have strong correlations with human-defined task characteristics, even when the decisions are learned spontaneously without no prior knowledge such as pre-computed task representations. 
We hope our work provide useful advice on training and interpreting multi-task models in NLP and we hope it will inspire future work in improving multi-task learning and cross-task generalization in NLP.

\section*{Limitations}
Although we have done much analysis on the correlation between learned routes and task characteristics, it is yet challenging to (1) ground each expert to human-understandable language skills; (2) understand their causal relationships. 
Much more needs to be discussed on how to systematically define the atomic/basic skills that are used in solving NLP tasks. 
In terms of model optimization, we find that we cannot achieve the best performance using the one-stage training strategy, and our best method takes more training time and needs more delicate hyper-parameters selection compared to the vanilla multi-task model. We hypothesize that there are optimization challenges in training task-level mixture-of-expert models. We hope future work can investigate and address this problem.

\section*{Acknowledgments}
We thank authors and crowd-workers of all datasets used in our study. We thank huggingface datasets team \cite{lhoest-etal-2021-datasets} for making NLP datasets more accessible. We thank anonymous reviewers, members of USC INK Lab and USC NLP community for their valuable feedback. This work is supported in part by the Office of the Director of National Intelligence (ODNI), Intelligence Advanced Research Projects Activity (IARPA), via Contract No. 2019-19051600007; the DARPA MCS program under Contract No. N660011924033; the Defense Advanced Research Projects Agency with award W911NF-19-20271; NSF IIS 2048211.

\bibliography{anthology,custom}
\bibliographystyle{acl_natbib}

\appendix

\section{Computing Task Representations}
In the following, we describe the method to construct the task representations used in \S\ref{ssec:design_choices}.
\label{app:design_choices}
\paragraph{TaskEmb.} Task2Vec \citep{Achille_2019_ICCV} is a method to generate tasks embedding for visual classification tasks based fisher information matrix (FIM). It was then extended to NLP domain \cite{vu-etal-2020-exploring, wang2021gradtask} and was found to be useful. We compute the empirical fisher and use them as task representations, following \citet{vu-etal-2020-exploring}. 
Specifically, given a model $P_{\theta}$ parameterized by $\theta$ (\textit{e.g.}, a BART-Base model) and a set of labeled examples $\{(x, y)\}$, we first fine-tune the model on the examples, then compute the fisher information matrix: 
\begin{equation}
\small
\label{eq:fisherAdapter}
{F_{\theta}=\frac{1}{n}\sum_{i=1}^{n}\left[\nabla_{\theta}\log{P}_{\theta}\left(y^{i}|x^{i}\right) \nabla_{\theta} \log{P}_{\theta}\left(y^{i}|x^{i}\right)^{T}\right]}
\end{equation}
To reduce the computational complexity, 
(1) we only use the diagonal entries of $F_{\theta}$, following \citet{Achille_2019_ICCV} and \citet{vu-etal-2020-exploring}; 
(2) we use a parameter-efficient fine-tuning method named adapter fine-tuning \cite{Houlsby2019ParameterEfficientTL} and only compute the FIM with respect to adapter parameters.
(3) we use PCA to reduce the dimension ($d=768$, which is the same as TextEmb), as we will use these representations as input to our router in the task-level MoE model.

\paragraph{TextEmb and FT-TextEmb.} For TextEmb, we first concatenate the input sequence $x$ and the output sequence $y$ into a longer sequence, and feed it to the encoder of BART to get token-level representations. For TextEmb-AVG, we compute the average over tokens for each example, and then average over all examples, to get a final vector as task representation. For TextEmb-BOS, we average the BOS representation of all examples\footnote{We later found out that this is less meaningful since BART pre-training does not train these BOS tokens with any special objective.}. For fair comparison with TaskEmb, which fine-tunes the model on labeled examples and thus may obtain extra information through this process, we also include FT-TextEmb-AVG and FT-TextEmb-BOS in our comparison. In these two variants, the BART model is first fine-tune on the labeled examples $\{(x, y)\}$.

\section{Additional Experiment Details}

\subsection{Multi-task Learning Experiments}
\label{app:multitask_details}
We concatenate the $D_{train}$ of the 120 tasks in $\mathcal{T}_{train}$ into a large dataset and use it for multi-task learning. We adopt heterogeneous batching \cite{aghajanyan-etal-2021-muppet}, \textit{i.e.}, each batch contains examples from different tasks. For the vanilla multi-task baseline, we train the model for 30,000 steps, with the batch size equals to 32 and the learning rate equals to 3e-5. For BART-Large we use the same setting, except that the learning rate is set to 1e-5. We use validation every 3,000 steps and select the best model based on validation performance. 

For the task-level MoE models, they are trained with a basic learning rate of 1e-5, while we set the router with bigger learning rate of 1e-3 based on our pilot experiments following \citet{ponti2022combining}. 
For the task representations, we use 1e-2 as learning rate when they are randomly initialized, and 1e-3 when initialized from pre-computed representations. 
We train the model for 60,000 steps because it takes more exploration time for the routes and experts to be stable. All models are trained with Adam optimizer~\citep{kingma2014adam}. 

\subsection{Few-shot Adaptation Experiments}
For few-shot fine-tuning we mainly follow the experiment setting in \citet{ye-etal-2021-crossfit}. Each task has five different few-shot samples of $(D_{train}, D_{dev})$. We train on $D_{train}$ for 1000 steps, and validate on $D_{dev}$ every 100 steps. We run a grid search for learning rate \{1e-5, 2e-5, 5e-5\} and batch size \{2,4,8\} for each few-shot sample. Finally, the model with best $D_{dev}$ performance is evaluated on $D_{test}$, the we report the performance on $D_{test}$.

\subsection{Zero-shot Experiments}
\label{app:zeroshot_setting}
\paragraph{Data.}
Following \citet{sanh2022multitask} and \citet{lin2022unsupervised}, we use the prompt templates in the Public Pool of Prompts (P3) \cite{bach-etal-2022-promptsource} to change texts from various NLP tasks into a unified text-to-text format.
To save compute, we use a sub-sampled version of P3 dataset. We use up to 5k examples for $D_{train}$, 1k examples for both $D_{dev}$ and $D_{test}$ following \citet{lin2022unsupervised} for all tasks.
We use 36 upstream tasks (which is the same as T0 upstream learning) for $\mathcal{T}_{train}$ and use 10 unseen tasks as our $\mathcal{T}_{test}$. $D_{train}$ for tasks in $\mathcal{T}_{train}$ are used for upstream learning; $D_{test}$ for tasks in $\mathcal{T}_{test}$ are used for reporting the performance. For simplicity, we only keep the prompt that can be evaluated with accuracy, and we report the mean acurracy for all tasks in $\mathcal{T}_{test}$.
\paragraph{Training.} (1) For Multi-task BART-Base and Random Task Routing (2/3), we use 1e-5 as the learning rate, 16 as training batch size, and the total training steps is set to be 200k. (2) For the \colorbox{red!10}{(c)}+\colorbox{blue!10}{(d)}+\colorbox{green!10}{(h)}+\colorbox{teal!10}{(l)}+\colorbox{brown!10}{(m)} model, we use 1e-5 as the base learning rate for experts and 1e-3 for the router. We train the model for 200k steps. (3) For the \colorbox{red!10}{(c)}+\colorbox{blue!10}{(d)} +\colorbox{green!10}{(h)}+\colorbox{teal!10}{(l)}+\colorbox{brown!10}{(n)} model, we use 1e-5 as the base learning rate for experts and 1e-3 for router. For the first learning stage we train for 60k steps, and 200k steps for the second stage. For both MoE models we use a batch size as 4. 
In this zero-shot setting, the task representation is computed by applying TextEmb-AVG \colorbox{green!10}{(h)} to the prompt templates.

\section{Extended Results and Analysis}

\subsection{Loss and Performance Discrepancy} 
\begin{figure}[t]
    \centering
    \includegraphics[width=0.5\textwidth]{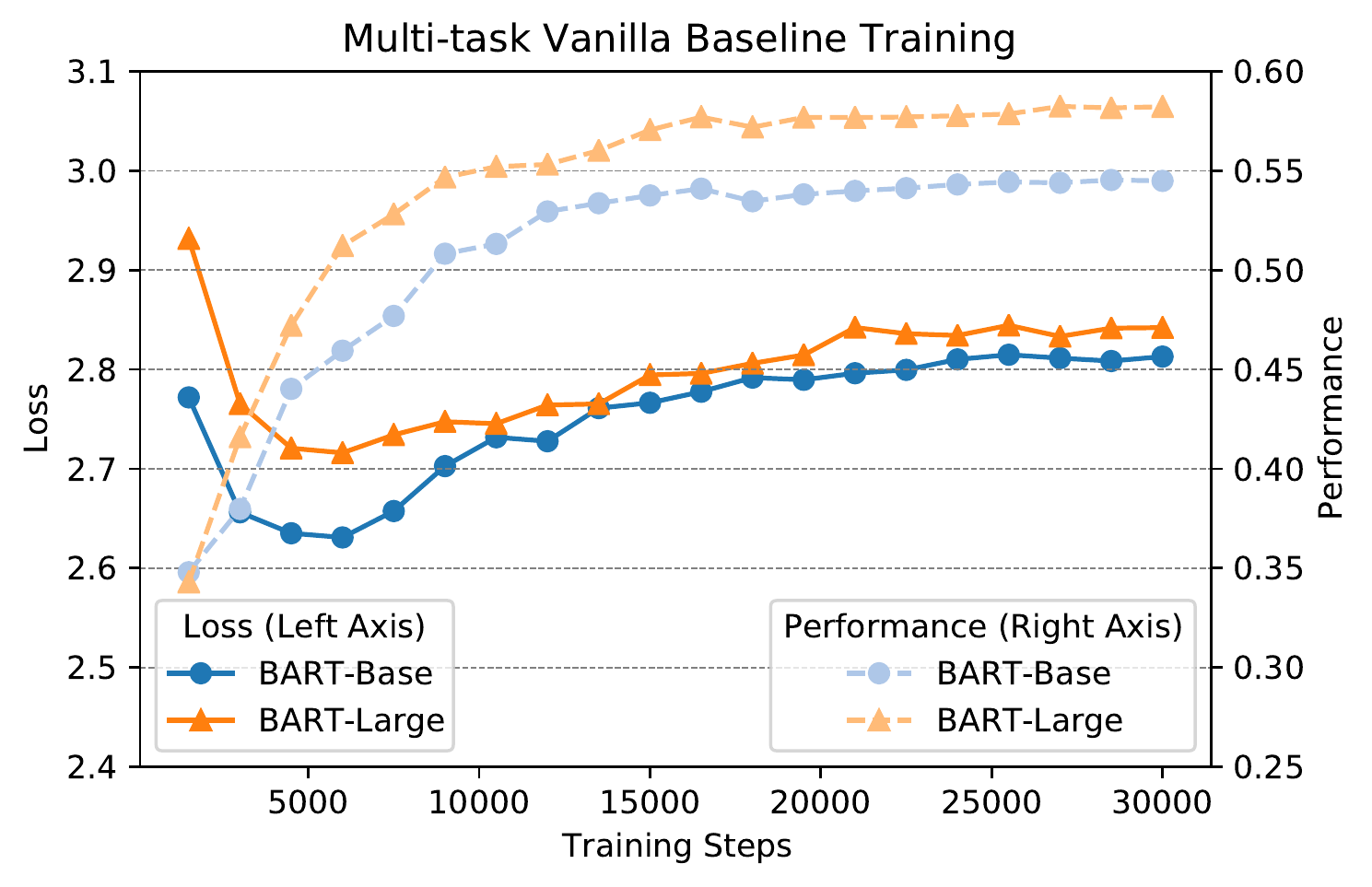}
    \caption{\textbf{Dev loss and dev performance discrepancy when training multi-task transformer baselines.} We found that smaller dev loss does not guarantee better dev performance. Dev losses tend to plunge then rise, while dev performance continue to increase. BART-Large outperforms BART-Base despite larger dev loss.}
    \label{fig:vanilla_training}
\end{figure}

In Fig.~\ref{fig:vanilla_training}, we plot the $D_{dev}$ loss and performance during multitask learning. We conclude that $D_{dev}$ loss does not align well with the final metrics, and thus validation should be done with the final metrics.
\label{app:discrepancy}

\subsection{Full Manual Feature Correlation Results}
\begin{figure*}[!ht]
    \centering
    \vspace{-1cm}
    \includegraphics[width=0.85\textwidth]{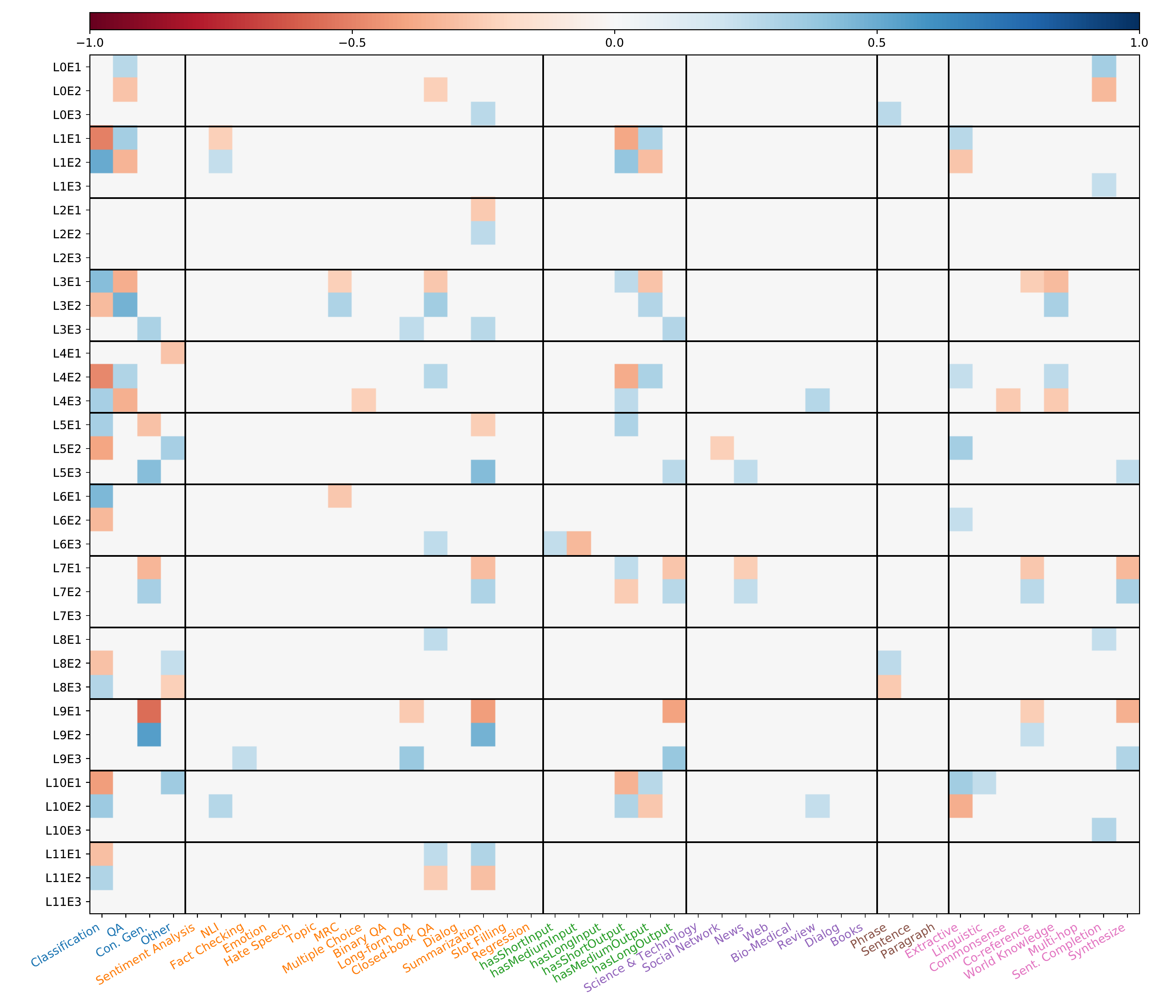}
    \vspace{-0.2cm}
    \caption{\textbf{Pearson Correlation Between Learned Routes and Manual Features.} Correlation with $p$<$0.01$ are visualized. The correlation is based on a \colorbox{red!10}{(c)}+\colorbox{blue!10}{(d)}+\colorbox{green!10}{(g)}+\colorbox{teal!10}{(k)} model.}
    \label{fig:corr-random-full}
    \vspace{0.2cm}
    \includegraphics[width=0.85\textwidth]{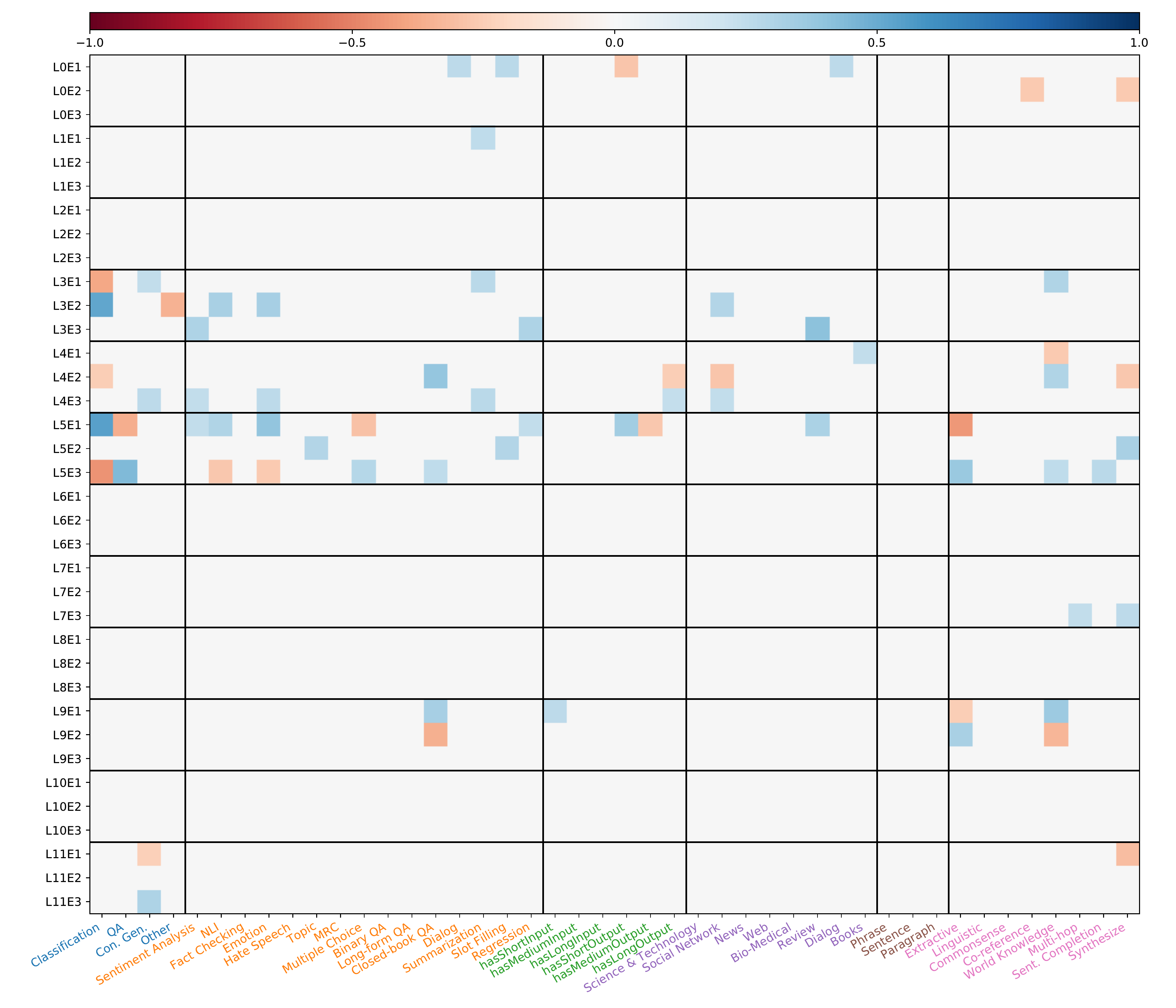}
    \vspace{-0.2cm}
    \caption{\textbf{Pearson Correlation Between Learned Routes and Manual Features.} Correlation with $p$<$0.01$ are visualized. The correlation is based on a \colorbox{red!10}{(c)}+\colorbox{blue!10}{(d)}+\colorbox{green!10}{(j)}+\colorbox{teal!10}{(k)} model.}
    \label{fig:corr-random-fisher}
\end{figure*}

We show the full results of Pearson Correlation between learned routes and manual features in Figure~\ref{fig:corr-random-full} and Figure~\ref{fig:corr-random-fisher}. Figure~\ref{fig:corr-random-full} is based on routes in the \colorbox{red!10}{(c)}+\colorbox{blue!10}{(d)}+\colorbox{green!10}{(g)}+\colorbox{teal!10}{(k)} model, and Figure~\ref{fig:corr-random-fisher} is based on the \colorbox{red!10}{(c)}+\colorbox{blue!10}{(d)}+\colorbox{green!10}{(j)}+\colorbox{teal!10}{(k)} model.

\subsection{Further Investigation on Selection Functions}
\label{app:softmax}
In our initial experiments, the implementation of softmax does not have temperature annealing. When we include this trick, the performance is comparable to gumbel-softmax ST.

\section{Discussion on Contemporary Works}\label{app:contemporary-works}
Training dynamical models that condition the computation on task information is a growing and active research field. Several contemporary works \cite{ponti2022combining, gupta2022sparsely,asai2022attempt} are studying this problem.
We share similar motivations with these works; meanwhile, these works and ours differ in methodology and research focus. We would like to highlight that (1) we conduct extensive analysis on interpreting the learned routes and experts in \S\ref{sec:interpret}; (2) we use 120 seen tasks and 18 unseen tasks, which is more diverse, and creates a challenging learning setting. We hope our findings are useful to the EMNLP community.

\newpage
\clearpage
\onecolumn
\section{Tasks Used and References}
\label{app:all_tasks}
We list all the tasks used in this paper in Table~\ref{tab:ontology} and its corresponding manual feature labels in Table ~\ref{tab:hand_features}.

\scriptsize
\begin{longtable}{lllll}
\caption{Tasks used in this work.}\label{tab:ontology}\\
\toprule
\textbf{Task Name} & \textbf{Ontology} & \textbf{Reference} \\
\midrule
\endfirsthead
\toprule
\textbf{Task Name} & \textbf{Ontology} & \textbf{Reference} \\
\midrule
\endhead
\bottomrule \multicolumn{4}{r}{Continued on next page} \\
\endfoot
\endlastfoot
acronym\_identification &	other &	\citealt{pouran-ben-veyseh-etal-2020-acronym}	\\
ade\_corpus\_v2-classification &	cls/other &	\citealt{GURULINGAPPA2012885}	\\
ade\_corpus\_v2-dosage &	other/slot filling & \citealt{GURULINGAPPA2012885}	\\
ade\_corpus\_v2-effect &	other/slot filling & \citealt{GURULINGAPPA2012885}	\\
adversarialqa &	qa/machine reading comprehension &	\citealt{bartolo-etal-2020-beat}	\\
aeslc &	cg/summarization & \citealt{zhang-tetreault-2019-email}	\\
ag\_news &	cls/topic &	\href{http://groups.di.unipi.it/~gulli/AG_corpus_of_news_articles.html}{Gulli (link)}	\\
ai2\_arc &	qa/multiple-choice qa &	\citealt{Clark2018ThinkYH}	\\
amazon\_polarity &	cls/sentiment analysis & \citealt{McAuley2013HiddenFA}	\\
anli &	cls/nli & \citealt{nie-etal-2020-adversarial}	\\
app\_reviews &	other/regression &	Missing\\ 
aqua\_rat &	qa/multiple-choice qa &	\citealt{ling-etal-2017-program}	\\
art (abductive nli) &	other &	\citealt{bhagavatula2020abductive}	\\
aslg\_pc12 &	other &	\citealt{Othman2012EnglishASLGP}	\\
biomrc &	qa/machine reading comprehension &	\citealt{pappas-etal-2020-biomrc}	\\
blimp-anaphor\_gender\_agreement &	other/linguistic phenomenon & \citealt{warstadt2019blimp}	\\
blimp-anaphor\_number\_agreement &	other/linguistic phenomenon & \citealt{warstadt2019blimp}	\\
blimp-determiner\_noun\_agreement\_with\_adj\_irregular\_1 &	other/linguistic phenomenon & \citealt{warstadt2019blimp}	\\
blimp-ellipsis\_n\_bar\_1 &	other/linguistic phenomenon & \citealt{warstadt2019blimp}	\\
blimp-ellipsis\_n\_bar\_2 &	other/linguistic phenomenon & \citealt{warstadt2019blimp}	\\
blimp-existential\_there\_quantifiers\_1 &	other/linguistic phenomenon & \citealt{warstadt2019blimp}	\\
blimp-irregular\_past\_participle\_adjectives &	other/linguistic phenomenon & \citealt{warstadt2019blimp}	\\
blimp-sentential\_negation\_npi\_licensor\_present &	other/linguistic phenomenon & \citealt{warstadt2019blimp}	\\
blimp-sentential\_negation\_npi\_scope &	other/linguistic phenomenon & \citealt{warstadt2019blimp}	\\
blimp-wh\_questions\_object\_gap &	other/linguistic phenomenon & \citealt{warstadt2019blimp}	\\
boolq &	qa/binary &	\citealt{clark-etal-2019-boolq}	\\
break-QDMR &	other &	\citealt{wolfson-etal-2020-break}	\\
break-QDMR-high-level &	other & \citealt{wolfson-etal-2020-break}	\\
circa &	cls/other &	\citealt{louis-etal-2020-id}	\\
climate\_fever &	cls/fact checking &	\citealt{Diggelmann2020CLIMATEFEVERAD}	\\
codah &	qa/multiple-choice qa &	\citealt{chen-etal-2019-codah}	\\
common\_gen &	other &	\citealt{lin-etal-2020-commongen}	\\
commonsense\_qa &	qa/multiple-choice qa &	\citealt{talmor-etal-2019-commonsenseqa}	\\
cos\_e &	other/generate explanation & \citealt{rajani-etal-2019-explain}	\\
cosmos\_qa &	qa/multiple-choice qa &	\citealt{huang-etal-2019-cosmos}	\\
crawl\_domain &	other &	\citealt{zhang-etal-2020-semi}	\\
crows\_pairs &	other &	\citealt{nangia-etal-2020-crows}	\\
dbpedia\_14 &	cls/topic &	\citealt{Lehmann2015DBpediaA}	\\
definite\_pronoun\_resolution &	other &	\citealt{rahman-ng-2012-resolving}	\\
discovery &	cls/other &	\citealt{sileo-etal-2019-mining}	\\
dream &	qa/multiple-choice qa &	\citealt{sun-etal-2019-dream}	\\
duorc &	qa/machine reading comprehension &	\citealt{saha-etal-2018-duorc}	\\
e2e\_nlg\_cleaned &	other &	\citealt{dusek.etal2020:csl, dusek-etal-2019-semantic}	\\
eli5-askh &	qa/long-form qa & \citealt{fan-etal-2019-eli5}	\\
eli5-asks &	qa/long-form qa & \citealt{fan-etal-2019-eli5}	\\
eli5-eli5 &	qa/long-form qa & \citealt{fan-etal-2019-eli5}	\\
emo &	cls/emotion & \citealt{chatterjee-etal-2019-semeval}	\\
emotion &	cls/emotion & \citealt{saravia-etal-2018-carer}	\\
empathetic\_dialogues &	cg/dialogue & \citealt{rashkin-etal-2019-towards}	\\
ethos-directed\_vs\_generalized &	cls/hate speech detection &	\citealt{Mollas2020ETHOSAO}	\\
ethos-disability &	cls/hate speech detection &	\citealt{Mollas2020ETHOSAO}	\\
ethos-gender &	cls/hate speech detection &	\citealt{Mollas2020ETHOSAO}	\\
ethos-national\_origin &	cls/hate speech detection &	\citealt{Mollas2020ETHOSAO}	\\
ethos-race &	cls/hate speech detection &	\citealt{Mollas2020ETHOSAO}	\\
ethos-religion &	cls/hate speech detection &	\citealt{Mollas2020ETHOSAO}	\\
ethos-sexual\_orientation &	cls/hate speech detection &	\citealt{Mollas2020ETHOSAO}	\\
financial\_phrasebank &	cls/sentiment analysis &	\citealt{financial-phrasebank}	\\
freebase\_qa &	qa/closed-book qa &	\citealt{jiang-etal-2019-freebaseqa}	\\
gigaword &	cg/summarization &	\citealt{napoles-etal-2012-annotated}	\\
glue-cola &	cls/other &	\citealt{warstadt-etal-2019-neural}	\\
glue-mnli &	cls/nli & \citealt{williams-etal-2018-broad}	\\
glue-mrpc &	cls/paraphrase & \citealt{dolan-brockett-2005-automatically}\\
glue-qnli &	cls/nli &	\citealt{rajpurkar-etal-2016-squad}	\\
glue-qqp &	cls/paraphrase & \href{http://data.quora.com/First-Quora-Dataset-Release-Question-Pairs}{(link)}	\\
glue-rte &	cls/nli &	\begin{tabular}[c]{@{}l@{}}\citealt{dagan2005pascal, bar2006second}\\\citealt{giampiccolo2007third, bentivogli2009fifth}\end{tabular}	\\
glue-sst2 &	cls/sentiment analysis &	\citealt{socher-etal-2013-recursive}	\\
glue-wnli &	cls/nli & \citealt{levesque2012winograd}	\\
google\_wellformed\_query &	cls/other &	\citealt{faruqui-das-2018-identifying}	\\
hate\_speech18 &	cls/hate speech detection &	\citealt{gibert2018hate}	\\
hate\_speech\_offensive &	cls/hate speech detection &	\citealt{hateoffensive}	\\
hatexplain &	cls/hate speech detection &	\citealt{mathew2020hatexplain}	\\
health\_fact &	cls/fact checking &	\citealt{kotonya-toni-2020-explainable-automated}	\\
hellaswag &	qa/multiple-choice qa &	\citealt{zellers-etal-2019-hellaswag}	\\
hotpot\_qa &	qa/machine reading comprehension &	\citealt{yang-etal-2018-hotpotqa}	\\
imdb &	cls/sentiment analysis & \citealt{maas-etal-2011-learning}	\\
jeopardy &	qa/closed-book qa &	\href{https://www.reddit.com/r/datasets/comments/1uyd0t/200000_jeopardy_questions_in_a_json_file/}{(link)}	\\
kilt\_ay2 &	other/entity linking &	\citealt{hoffart-etal-2011-robust}	\\
kilt\_fever &	cls/fact checking &	\citealt{thorne-etal-2018-fever}	\\
kilt\_hotpotqa &	qa/closed-book qa &	\citealt{yang-etal-2018-hotpotqa}	\\
kilt\_nq &	qa/closed-book qa &	\citealt{kwiatkowski-etal-2019-natural}	\\
kilt\_trex &	qa/closed-book qa &	\citealt{elsahar-etal-2018-rex}	\\
kilt\_wow &	cg/dialogue &	\citealt{dinan2018wizard}	\\
kilt\_zsre &	qa/closed-book qa &	\citealt{levy-etal-2017-zero}	\\
lama-conceptnet &	qa/closed-book qa &	\citealt{petroni-etal-2019-language,petroni2020how}	\\
lama-google\_re &	qa/closed-book qa &	\citealt{petroni-etal-2019-language,petroni2020how}	\\
lama-squad &	qa/closed-book qa &	\citealt{petroni-etal-2019-language,petroni2020how}	\\
lama-trex &	qa/closed-book qa &	\citealt{petroni-etal-2019-language,petroni2020how}	\\
liar &	cls/fact checking &	\citealt{wang-2017-liar}	\\
limit &	other &	\citealt{manotas-etal-2020-limit}	\\
math\_qa &	qa/multiple-choice qa &	\citealt{amini-etal-2019-mathqa}	\\
mc\_taco &	qa/binary &	\citealt{zhou-etal-2019-going}	\\
medical\_questions\_pairs &	cls/paraphrase & \citealt{medical-qqp}	\\
mocha &	other/regression &	\citealt{chen-etal-2020-mocha}	\\
multi\_news &	cg/summarization & \citealt{fabbri-etal-2019-multi}	\\
numer\_sense &	qa/closed-book qa &	\citealt{lin-etal-2020-birds}	\\
onestop\_english &	cls/other &	\citealt{vajjala-lucic-2018-onestopenglish}	\\
openbookqa &	qa/multiple-choice qa &	\citealt{mihaylov-etal-2018-suit}	\\
paws &	cls/paraphrase & \citealt{zhang-etal-2019-paws}	\\
piqa &	other &	\citealt{Bisk2020}	\\
poem\_sentiment &	cls/sentiment analysis & \citealt{sheng-uthus-2020-investigating}	\\
proto\_qa &	other &	\citealt{boratko-etal-2020-protoqa}	\\
qa\_srl &	other &	\citealt{he-etal-2015-question}	\\
qasc &	qa/multiple-choice qa &	\citealt{Khot_Clark_Guerquin_Jansen_Sabharwal_2020}	\\
quail &	qa/multiple-choice qa &	\citealt{Rogers_Kovaleva_Downey_Rumshisky_2020}	\\
quarel &	qa/multiple-choice qa &	\citealt{Tafjord_Clark_Gardner_Yih_Sabharwal_2019}	\\
quartz-no\_knowledge &	qa/multiple-choice qa &	\citealt{tafjord-etal-2019-quartz}	\\
quartz-with\_knowledge &	qa/multiple-choice qa &	\citealt{tafjord-etal-2019-quartz}	\\
quoref &	qa/machine reading comprehension &	\citealt{dasigi-etal-2019-quoref}	\\
race-high &	qa/multiple-choice qa &	\citealt{lai-etal-2017-race}	\\
race-middle &	qa/multiple-choice qa &	\citealt{lai-etal-2017-race}	\\
reddit\_tifu-title &	cg/summarization &	\citealt{kim-etal-2019-abstractive}	\\
reddit\_tifu-tldr &	cg/summarization &	\citealt{kim-etal-2019-abstractive}	\\
ropes &	qa/machine reading comprehension &	\citealt{lin-etal-2019-reasoning}	\\
rotten\_tomatoes &	cls/sentiment analysis & \citealt{pang-lee-2005-seeing}	\\
samsum &	cg/summarization &	\citealt{gliwa-etal-2019-samsum}	\\
scicite &	cls/other &	\citealt{cohan-etal-2019-structural}	\\
sciq &	qa/multiple-choice qa &	\citealt{welbl-etal-2017-crowdsourcing}	\\
scitail &	cls/nli & \citealt{scitail}	\\
search\_qa &	qa/closed-book qa &	\citealt{Dunn2017SearchQAAN}	\\
sick &	cls/nli &	\citealt{marelli-etal-2014-sick}	\\
sms\_spam &	cls/other &	\citealt{sms_spam}	\\
social\_i\_qa &	qa/multiple-choice qa &	\citealt{sap-etal-2019-social}	\\
spider &	cg/other &	\citealt{yu-etal-2018-spider}	\\
squad-no\_context &	qa/closed-book qa &	\citealt{rajpurkar-etal-2016-squad}	\\
squad-with\_context &	qa/machine reading comprehension &	\citealt{rajpurkar-etal-2016-squad}	\\
superglue-cb &	cls/nli & \citealt{Marneffe_Simons_Tonhauser_2019}	\\
superglue-copa &	qa/multiple-choice qa &	\citealt{gordon-etal-2012-semeval} \\
superglue-multirc &	qa/multiple-choice qa &	\citealt{khashabi-etal-2018-looking}	\\
superglue-record &	qa/machine reading comprehension & \citealt{Zhang2018ReCoRDBT}	\\
superglue-rte &	cls/nli & \begin{tabular}[c]{@{}l@{}}\citealt{dagan2005pascal, bar2006second}\\\citealt{giampiccolo2007third, bentivogli2009fifth}\end{tabular}	\\
superglue-wic &	cls/other &	\citealt{pilehvar-camacho-collados-2019-wic}	\\
superglue-wsc &	cls/other &	\citealt{levesque2012winograd}	\\
swag &	qa/multiple-choice qa &	\citealt{zellers-etal-2018-swag}	\\
tab\_fact &	cls/fact checking &	\citealt{Chen2020TabFact}	\\
trec &	cls/other &	\citealt{li-roth-2002-learning,hovy-etal-2001-toward}	\\
trec-finegrained &	cls/other &	\citealt{li-roth-2002-learning,hovy-etal-2001-toward}	\\
tweet\_eval-emoji &	cls/emotion & \citealt{barbieri-etal-2020-tweeteval}	\\
tweet\_eval-emotion &	cls/emotion &	\citealt{barbieri-etal-2020-tweeteval}	\\
tweet\_eval-hate &	cls/emotion &	\citealt{barbieri-etal-2020-tweeteval}	\\
tweet\_eval-irony &	cls/emotion &	\citealt{barbieri-etal-2020-tweeteval}	\\
tweet\_eval-offensive &	cls/emotion &	\citealt{barbieri-etal-2020-tweeteval}	\\
tweet\_eval-sentiment &	cls/emotion &	\citealt{barbieri-etal-2020-tweeteval}	\\
tweet\_eval-stance\_abortion &	cls/emotion &	\citealt{barbieri-etal-2020-tweeteval}	\\
tweet\_eval-stance\_atheism &	cls/emotion &	\citealt{barbieri-etal-2020-tweeteval}	\\
tweet\_eval-stance\_climate &	cls/emotion &	\citealt{barbieri-etal-2020-tweeteval}	\\
tweet\_eval-stance\_feminist &	cls/emotion &	\citealt{barbieri-etal-2020-tweeteval}	\\
tweet\_eval-stance\_hillary &	cls/emotion &	\citealt{barbieri-etal-2020-tweeteval}	\\
tweet\_qa &	qa/machine reading comprehension &	\citealt{xiong-etal-2019-tweetqa}	\\
web\_questions &	qa/closed-book qa &	\citealt{berant-etal-2013-semantic}	\\
wiki\_auto &	cls/other &	\citealt{jiang-etal-2020-neural}	\\
wiki\_bio &	cg/other &	\citealt{lebret-etal-2016-neural}	\\
wiki\_qa &	cls/other &	\citealt{yang-etal-2015-wikiqa}	\\
wiki\_split &	cg/other & \citealt{botha-etal-2018-learning}	\\
wikisql &	cg/other &	\citealt{zhongSeq2SQL2017}	\\
wino\_grande &	qa/multiple-choice qa &	\citealt{Sakaguchi_Le_Bras_Bhagavatula_Choi_2020}	\\
wiqa &	qa/multiple-choice qa &	\citealt{tandon-etal-2019-wiqa}	\\
xsum &	cg/summarization &	\citealt{narayan-etal-2018-dont}	\\
yahoo\_answers\_topics &	cls/topic &	\href{https://webscope.sandbox.yahoo.com/catalog.php?datatype=l}{(link)}	\\
yelp\_polarity &	cls/sentiment analysis & \citealt{zhang2015character}; \href{https://www.yelp.com/dataset}{(link)}	\\
yelp\_review\_full &	other/regression &	\citealt{zhang2015character}; \href{https://www.yelp.com/dataset}{(link)}	\\
cnn\_dailymail &  cg/summarization  & \citealt{nallapati-etal-2016-abstractive} \\
wiki\_hop & qa/multiple-choice qa & \citealt{welbl-etal-2018-constructing} \\
\bottomrule
\end{longtable}
\normalsize

\section{Random Task Partition}
\label{app:random_task_partition}
Different from the original random task partition used in \citet{ye-etal-2021-crossfit}, we remove yelp\_polarity and freebase\_qa from $\mathcal{T}_{test}$ because we observe unusual instability when doing few-shot fine-tuning on these tasks.
\begin{lstlisting}[language=json,firstnumber=1]
{
    "train": ['glue-mrpc', 'math_qa', 'quarel', 'e2e_nlg_cleaned', 'tweet_eval-stance_atheism', 'lama-squad', 'tab_fact', 'aqua_rat', 'tweet_eval-emoji', 'glue-wnli', 'codah', 'tweet_eval-offensive', 'wiki_qa', 'blimp-ellipsis_n_bar_1', 'openbookqa', 'sms_spam', 'acronym_identification', 'blimp-determiner_noun_agreement_with_adj_irregular_1', 'ethos-national_origin', 'spider', 'definite_pronoun_resolution', 'hellaswag', 'superglue-wsc', 'numer_sense', 'ade_corpus_v2-dosage', 'blimp-ellipsis_n_bar_2', 'kilt_ay2', 'squad-no_context', 'google_wellformed_query', 'xsum', 'wiqa', 'tweet_eval-stance_abortion', 'reddit_tifu-tldr', 'ade_corpus_v2-effect', 'qa_srl', 'ethos-religion', 'commonsense_qa', 'jeopardy', 'biomrc', 'superglue-multirc', 'ethos-race', 'eli5-askh', 'glue-qqp', 'paws', 'ethos-directed_vs_generalized', 'glue-sst2', 'mocha', 'tweet_eval-hate', 'glue-rte', 'blimp-anaphor_number_agreement', 'lama-conceptnet', 'hate_speech_offensive', 'superglue-wic', 'boolq', 'kilt_hotpotqa', 'quartz-no_knowledge', 'aslg_pc12', 'sick', 'tweet_eval-stance_climate', 'tweet_eval-sentiment', 'crows_pairs', 'glue-mnli', 'medical_questions_pairs', 'break-QDMR-high-level', 'qasc', 'imdb', 'ethos-gender', 'trec-finegrained', 'adversarialqa', 'onestop_english', 'web_questions', 'duorc', 'yelp_review_full', 'swag', 'proto_qa', 'scitail', 'tweet_eval-stance_feminist', 'limit', 'common_gen', 'scicite', 'blimp-irregular_past_participle_adjectives', 'social_i_qa', 'anli', 'kilt_zsre', 'cosmos_qa', 'superglue-record', 'squad-with_context', 'emotion', 'blimp-existential_there_quantifiers_1', 'race-middle', 'kilt_wow', 'sciq', 'wino_grande', 'rotten_tomatoes', 'superglue-cb', 'poem_sentiment', 'ropes', 'reddit_tifu-title', 'piqa', 'climate_fever', 'lama-google_re', 'search_qa', 'wiki_auto', 'mc_taco', 'blimp-wh_questions_object_gap', 'hotpot_qa', 'emo', 'kilt_nq', 'kilt_trex', 'quartz-with_knowledge', 'dbpedia_14', 'yahoo_answers_topics', 'app_reviews', 'superglue-copa', 'blimp-anaphor_gender_agreement', 'hate_speech18', 'gigaword', 'multi_news', 'aeslc', 'quail'],
    "dev": ['cos_e', 'kilt_fever', 'eli5-asks', 'trec', 'eli5-eli5', 'art', 'empathetic_dialogues', 'tweet_qa', 'wikisql', 'lama-trex', 'tweet_eval-stance_hillary', 'discovery', 'tweet_eval-emotion', 'liar', 'wiki_bio', 'dream', 'ade_corpus_v2-classification', 'health_fact', 'samsum', 'financial_phrasebank'],
    "test": ['quoref', 'wiki_split', 'ethos-disability', 'superglue-rte', 'glue-cola', 'ethos-sexual_orientation', 'blimp-sentential_negation_npi_scope', 'ai2_arc', 'amazon_polarity', 'race-high', 'blimp-sentential_negation_npi_licensor_present', 'tweet_eval-irony', 'break-QDMR', 'crawl_domain', 'glue-qnli', 'hatexplain', 'ag_news', 'circa'],
}
\end{lstlisting}
\section{Manually-Defined Features}
\scriptsize


\newcommand*\rot{\multicolumn{1}{R{90}{1em}}}
\begin{table}[thp]
\setlength\LTleft{-2in}
\setlength\LTright{-2in}
    \centering
    \resizebox{1.02\columnwidth}{!}{
    \begin{tabular}{lllllllllllllllllllll}
    \Xhline{3\arrayrulewidth}
      \textbf{Task Name} & \textbf{Science Technology} & \textbf{Social Network} & \textbf{News} & \textbf{Web} & \textbf{Bio-Medical} & \textbf{Review} & \textbf{Dialog} & \textbf{Books} & \textbf{Financial} & \textbf{Phrase} & \textbf{Sentence} & \textbf{Paragraph} & \textbf{Extractive} & \textbf{Linguistic} & \textbf{Commonsense} & \textbf{Co-reference} & \textbf{World Knowledge} & \textbf{Multi-hop} & \textbf{Sentence Completion} & \textbf{Synthesize} \\
       acronym\_identification & 1 & 0 & 0 & 0 & 0 & 0 & 0 & 0 & 0 & 1 & 0 & 0 & 1 & 1 & 0 & 1 & 0 & 0 & 0 & 0 \\ 
ade\_corpus\_v2-classification & 0 & 0 & 0 & 0 & 1 & 0 & 0 & 0 & 0 & 0 & 1 & 0 & 0 & 0 & 0 & 0 & 0 & 0 & 0 & 0 \\
ade\_corpus\_v2-dosage & 0 & 0 & 0 & 0 & 1 & 0 & 0 & 0 & 0 & 0 & 1 & 0 & 1 & 0 & 0 & 0 & 0 & 0 & 0 & 0 \\
ade\_corpus\_v2-effect & 0 & 0 & 0 & 0 & 1 & 0 & 0 & 0 & 0 & 0 & 1 & 0 & 1 & 0 & 0 & 0 & 0 & 0 & 0 & 0 \\
adversarialqa & 0 & 0 & 0 & 1 & 0 & 0 & 0 & 0 & 0 & 0 & 0 & 1 & 1 & 0 & 0 & 1 & 0 & 1 & 0 & 0 \\
aeslc & 0 & 1 & 0 & 0 & 0 & 0 & 0 & 0 & 0 & 0 & 0 & 1 & 0 & 0 & 0 & 1 & 0 & 0 & 0 & 1 \\
ag\_news & 0 & 0 & 1 & 0 & 0 & 0 & 0 & 0 & 0 & 0 & 1 & 0 & 0 & 0 & 0 & 0 & 0 & 0 & 0 & 0 \\
ai2\_arc & 1 & 0 & 0 & 0 & 0 & 0 & 0 & 0 & 0 & 0 & 0 & 1 & 0 & 0 & 1 & 0 & 1 & 0 & 0 & 0 \\
amazon\_polarity & 0 & 0 & 0 & 0 & 0 & 1 & 0 & 0 & 0 & 0 & 0 & 1 & 0 & 0 & 0 & 0 & 0 & 0 & 0 & 0 \\
anli & 0 & 0 & 0 & 1 & 0 & 0 & 0 & 0 & 0 & 0 & 0 & 1 & 0 & 0 & 0 & 0 & 0 & 0 & 0 & 0 \\
app\_reviews & 1 & 0 & 0 & 0 & 0 & 1 & 0 & 0 & 0 & 0 & 0 & 1 & 0 & 0 & 0 & 0 & 0 & 0 & 0 & 0 \\
aqua\_rat & 1 & 0 & 0 & 0 & 0 & 0 & 0 & 0 & 0 & 0 & 0 & 1 & 0 & 0 & 0 & 0 & 0 & 1 & 0 & 0 \\
art & 0 & 0 & 0 & 0 & 0 & 0 & 0 & 0 & 0 & 0 & 1 & 0 & 1 & 0 & 1 & 0 & 0 & 0 & 0 & 0 \\
aslg\_pc12 & 0 & 0 & 0 & 0 & 0 & 0 & 0 & 0 & 0 & 0 & 1 & 0 & 0 & 1 & 0 & 0 & 0 & 0 & 0 & 1 \\
biomrc & 0 & 0 & 0 & 0 & 1 & 0 & 0 & 0 & 0 & 0 & 0 & 1 & 1 & 0 & 0 & 0 & 0 & 0 & 1 & 0 \\
blimp-anaphor\_gender\_agreement & 0 & 0 & 0 & 0 & 0 & 0 & 0 & 0 & 0 & 0 & 1 & 0 & 1 & 1 & 1 & 0 & 0 & 0 & 0 & 0 \\
blimp-anaphor\_number\_agreement & 0 & 0 & 0 & 0 & 0 & 0 & 0 & 0 & 0 & 0 & 1 & 0 & 1 & 1 & 0 & 0 & 0 & 0 & 0 & 0 \\
blimp-determiner\_noun\_agreement\_with\_adj\_irregular\_1 & 0 & 0 & 0 & 0 & 0 & 0 & 0 & 0 & 0 & 0 & 1 & 0 & 1 & 1 & 0 & 0 & 0 & 0 & 0 & 0 \\
blimp-ellipsis\_n\_bar\_1 & 0 & 0 & 0 & 0 & 0 & 0 & 0 & 0 & 0 & 0 & 1 & 0 & 1 & 1 & 0 & 0 & 0 & 0 & 0 & 0 \\
blimp-ellipsis\_n\_bar\_2 & 0 & 0 & 0 & 0 & 0 & 0 & 0 & 0 & 0 & 0 & 1 & 0 & 1 & 1 & 0 & 0 & 0 & 0 & 0 & 0 \\
blimp-existential\_there\_quantifiers\_1 & 0 & 0 & 0 & 0 & 0 & 0 & 0 & 0 & 0 & 0 & 1 & 0 & 1 & 1 & 0 & 0 & 0 & 0 & 0 & 0 \\
blimp-irregular\_past\_participle\_adjectives & 0 & 0 & 0 & 0 & 0 & 0 & 0 & 0 & 0 & 0 & 1 & 0 & 1 & 1 & 0 & 0 & 0 & 0 & 0 & 0 \\
blimp-sentential\_negation\_npi\_licensor\_present & 0 & 0 & 0 & 0 & 0 & 0 & 0 & 0 & 0 & 0 & 1 & 0 & 1 & 1 & 0 & 0 & 0 & 0 & 0 & 0 \\
blimp-sentential\_negation\_npi\_scope & 0 & 0 & 0 & 0 & 0 & 0 & 0 & 0 & 0 & 0 & 1 & 0 & 1 & 1 & 0 & 0 & 0 & 0 & 0 & 0 \\
blimp-wh\_questions\_object\_gap & 0 & 0 & 0 & 0 & 0 & 0 & 0 & 0 & 0 & 0 & 1 & 0 & 1 & 1 & 0 & 0 & 0 & 0 & 0 & 0 \\
boolq & 0 & 0 & 0 & 1 & 0 & 0 & 0 & 0 & 0 & 0 & 0 & 1 & 0 & 0 & 0 & 0 & 1 & 1 & 0 & 0 \\
break-QDMR & 0 & 0 & 0 & 0 & 0 & 0 & 0 & 0 & 0 & 0 & 1 & 0 & 0 & 0 & 0 & 0 & 0 & 1 & 0 & 1 \\
break-QDMR-high-level & 0 & 0 & 0 & 0 & 0 & 0 & 0 & 0 & 0 & 0 & 1 & 0 & 0 & 0 & 0 & 0 & 0 & 1 & 0 & 1 \\
circa & 0 & 0 & 0 & 0 & 0 & 0 & 1 & 0 & 0 & 0 & 1 & 0 & 0 & 0 & 1 & 0 & 0 & 0 & 0 & 0 \\
climate\_fever & 0 & 0 & 0 & 1 & 0 & 0 & 0 & 0 & 0 & 0 & 1 & 0 & 0 & 0 & 1 & 0 & 1 & 0 & 0 & 0 \\
codah & 0 & 0 & 0 & 0 & 0 & 0 & 0 & 0 & 0 & 0 & 1 & 0 & 1 & 0 & 1 & 0 & 0 & 0 & 1 & 0 \\
common\_gen & 0 & 0 & 0 & 0 & 0 & 0 & 0 & 0 & 0 & 0 & 1 & 0 & 0 & 0 & 1 & 0 & 0 & 0 & 0 & 1 \\
commonsense\_qa & 0 & 0 & 0 & 0 & 0 & 0 & 0 & 0 & 0 & 0 & 1 & 0 & 1 & 0 & 1 & 0 & 0 & 0 & 0 & 0 \\
cos\_e & 0 & 0 & 0 & 0 & 0 & 0 & 0 & 0 & 0 & 0 & 1 & 0 & 0 & 0 & 1 & 0 & 0 & 0 & 0 & 0 \\
cosmos\_qa & 0 & 1 & 0 & 1 & 0 & 0 & 0 & 0 & 0 & 0 & 0 & 1 & 1 & 0 & 1 & 0 & 0 & 0 & 0 & 0 \\
crawl\_domain & 0 & 0 & 0 & 0 & 0 & 0 & 0 & 0 & 0 & 1 & 0 & 0 & 0 & 1 & 0 & 0 & 0 & 0 & 0 & 1 \\
crows\_pairs & 0 & 0 & 0 & 0 & 0 & 0 & 0 & 0 & 0 & 0 & 1 & 0 & 1 & 0 & 1 & 0 & 0 & 0 & 0 & 0 \\
dbpedia\_14 & 0 & 0 & 0 & 1 & 0 & 0 & 0 & 0 & 0 & 0 & 0 & 1 & 0 & 0 & 0 & 0 & 0 & 0 & 0 & 0 \\
definite\_pronoun\_resolution & 0 & 0 & 0 & 0 & 0 & 0 & 0 & 0 & 0 & 0 & 1 & 0 & 1 & 1 & 0 & 1 & 0 & 0 & 0 & 0 \\
discovery & 0 & 0 & 0 & 0 & 0 & 0 & 0 & 0 & 0 & 0 & 1 & 0 & 0 & 1 & 0 & 0 & 0 & 0 & 1 & 0 \\
dream & 0 & 0 & 0 & 0 & 0 & 0 & 1 & 0 & 0 & 0 & 0 & 1 & 1 & 0 & 1 & 0 & 0 & 0 & 0 & 0 \\
duorc & 0 & 0 & 0 & 1 & 0 & 0 & 0 & 0 & 0 & 0 & 0 & 1 & 1 & 0 & 1 & 0 & 0 & 1 & 0 & 0 \\
e2e\_nlg\_cleaned & 0 & 0 & 0 & 0 & 0 & 0 & 0 & 0 & 0 & 0 & 0 & 0 & 0 & 0 & 0 & 0 & 0 & 0 & 0 & 1 \\
eli5-askh & 0 & 1 & 0 & 1 & 0 & 0 & 0 & 0 & 0 & 0 & 1 & 0 & 0 & 0 & 0 & 1 & 0 & 0 & 0 & 0 \\
eli5-asks & 0 & 1 & 0 & 1 & 0 & 0 & 0 & 0 & 0 & 0 & 1 & 0 & 0 & 0 & 0 & 1 & 0 & 0 & 0 & 0 \\
eli5-eli5 & 0 & 1 & 0 & 1 & 0 & 0 & 0 & 0 & 0 & 0 & 1 & 0 & 0 & 0 & 0 & 1 & 1 & 0 & 0 & 0 \\
emo & 0 & 0 & 0 & 0 & 0 & 0 & 1 & 0 & 0 & 0 & 1 & 0 & 0 & 0 & 0 & 0 & 0 & 0 & 0 & 0 \\
emotion & 0 & 1 & 0 & 0 & 0 & 0 & 0 & 0 & 0 & 0 & 1 & 0 & 0 & 0 & 0 & 0 & 0 & 0 & 0 & 0 \\
empathetic\_dialogues & 0 & 0 & 0 & 0 & 0 & 0 & 1 & 0 & 0 & 0 & 1 & 0 & 0 & 0 & 0 & 0 & 0 & 0 & 0 & 0 \\
ethos-directed\_vs\_generalized & 0 & 1 & 0 & 0 & 0 & 0 & 0 & 0 & 0 & 0 & 1 & 0 & 0 & 0 & 0 & 0 & 0 & 0 & 0 & 0 \\
ethos-disability & 0 & 1 & 0 & 0 & 0 & 0 & 0 & 0 & 0 & 0 & 1 & 0 & 0 & 0 & 0 & 0 & 0 & 0 & 0 & 0 \\
ethos-gender & 0 & 1 & 0 & 0 & 0 & 0 & 0 & 0 & 0 & 0 & 1 & 0 & 0 & 0 & 0 & 0 & 0 & 0 & 0 & 0 \\
ethos-national\_origin & 0 & 1 & 0 & 0 & 0 & 0 & 0 & 0 & 0 & 0 & 1 & 0 & 0 & 0 & 0 & 0 & 0 & 0 & 0 & 0 \\
ethos-race & 0 & 1 & 0 & 0 & 0 & 0 & 0 & 0 & 0 & 0 & 1 & 0 & 0 & 0 & 0 & 0 & 0 & 0 & 0 & 0 \\
ethos-religion & 0 & 1 & 0 & 0 & 0 & 0 & 0 & 0 & 0 & 0 & 1 & 0 & 0 & 0 & 0 & 0 & 0 & 0 & 0 & 0 \\
ethos-sexual\_orientation & 0 & 1 & 0 & 0 & 0 & 0 & 0 & 0 & 0 & 0 & 1 & 0 & 0 & 0 & 0 & 0 & 0 & 0 & 0 & 0 \\
financial\_phrasebank & 0 & 0 & 0 & 0 & 0 & 0 & 0 & 0 & 1 & 0 & 1 & 0 & 0 & 0 & 0 & 0 & 0 & 0 & 0 & 0 \\
freebase\_qa & 0 & 0 & 0 & 0 & 0 & 0 & 0 & 0 & 0 & 0 & 1 & 0 & 0 & 0 & 0 & 0 & 1 & 0 & 0 & 0 \\
gigaword & 0 & 0 & 1 & 0 & 0 & 0 & 0 & 0 & 0 & 0 & 1 & 0 & 0 & 0 & 0 & 0 & 0 & 0 & 0 & 0 \\
glue-cola & 0 & 0 & 0 & 0 & 0 & 0 & 0 & 0 & 0 & 0 & 1 & 0 & 0 & 1 & 0 & 0 & 0 & 0 & 0 & 0 \\
glue-mnli & 0 & 0 & 0 & 1 & 0 & 0 & 1 & 1 & 0 & 0 & 1 & 0 & 0 & 0 & 0 & 0 & 0 & 0 & 0 & 0 \\
glue-mrpc & 0 & 0 & 1 & 0 & 0 & 0 & 0 & 0 & 0 & 0 & 1 & 0 & 0 & 1 & 0 & 0 & 0 & 0 & 0 & 0 \\
glue-qnli & 0 & 0 & 0 & 1 & 0 & 0 & 0 & 0 & 0 & 0 & 1 & 0 & 0 & 0 & 0 & 0 & 0 & 0 & 0 & 0 \\
glue-qqp & 0 & 1 & 0 & 1 & 0 & 0 & 0 & 0 & 0 & 0 & 1 & 0 & 0 & 1 & 0 & 0 & 0 & 0 & 0 & 0 \\
glue-rte & 0 & 0 & 1 & 1 & 0 & 0 & 0 & 0 & 0 & 0 & 1 & 0 & 0 & 0 & 0 & 0 & 0 & 0 & 0 & 0 \\
glue-sst2 & 0 & 0 & 0 & 0 & 0 & 1 & 0 & 0 & 0 & 0 & 1 & 0 & 0 & 0 & 0 & 0 & 0 & 0 & 0 & 0 \\
glue-wnli & 0 & 0 & 0 & 0 & 0 & 0 & 0 & 1 & 0 & 0 & 1 & 0 & 0 & 0 & 0 & 0 & 0 & 0 & 0 & 0 \\
google\_wellformed\_query & 0 & 0 & 0 & 1 & 0 & 0 & 0 & 0 & 0 & 0 & 1 & 0 & 0 & 1 & 0 & 0 & 0 & 0 & 0 & 0 \\
hate\_speech18 & 0 & 1 & 0 & 0 & 0 & 0 & 0 & 0 & 0 & 0 & 1 & 0 & 0 & 0 & 0 & 0 & 0 & 0 & 0 & 0 \\
hate\_speech\_offensive & 0 & 1 & 0 & 0 & 0 & 0 & 0 & 0 & 0 & 0 & 1 & 0 & 0 & 0 & 0 & 0 & 0 & 0 & 0 & 0 \\
hatexplain & 0 & 1 & 0 & 0 & 0 & 0 & 0 & 0 & 0 & 0 & 1 & 0 & 0 & 0 & 0 & 0 & 0 & 0 & 0 & 0 \\
health\_fact & 0 & 0 & 1 & 0 & 0 & 0 & 0 & 0 & 0 & 0 & 1 & 0 & 0 & 0 & 0 & 0 & 1 & 0 & 0 & 0 \\
hellaswag & 0 & 0 & 0 & 0 & 0 & 0 & 0 & 0 & 0 & 0 & 1 & 0 & 1 & 0 & 1 & 0 & 0 & 0 & 1 & 0 \\
hotpot\_qa & 0 & 0 & 0 & 1 & 0 & 0 & 0 & 0 & 0 & 0 & 0 & 1 & 1 & 0 & 0 & 0 & 0 & 1 & 0 & 0 \\
imdb & 0 & 0 & 0 & 0 & 0 & 1 & 0 & 0 & 0 & 0 & 0 & 1 & 0 & 0 & 0 & 0 & 0 & 0 & 0 & 0 \\
jeopardy & 0 & 0 & 0 & 1 & 0 & 0 & 0 & 0 & 0 & 0 & 1 & 0 & 0 & 0 & 0 & 0 & 1 & 0 & 0 & 0 \\
kilt\_ay2 & 0 & 0 & 0 & 1 & 0 & 0 & 0 & 0 & 0 & 1 & 0 & 0 & 0 & 0 & 0 & 0 & 1 & 0 & 0 & 0 \\
kilt\_fever & 0 & 0 & 0 & 0 & 0 & 0 & 0 & 0 & 0 & 0 & 1 & 0 & 0 & 0 & 1 & 0 & 1 & 0 & 0 & 0 \\
kilt\_hotpotqa & 0 & 0 & 0 & 0 & 0 & 0 & 0 & 0 & 0 & 0 & 1 & 0 & 0 & 0 & 0 & 0 & 1 & 1 & 0 & 0 \\
kilt\_nq & 0 & 0 & 0 & 0 & 0 & 0 & 0 & 0 & 0 & 0 & 1 & 0 & 0 & 0 & 0 & 0 & 1 & 1 & 0 & 0 \\
kilt\_trex & 0 & 0 & 0 & 0 & 0 & 0 & 0 & 0 & 0 & 1 & 0 & 0 & 0 & 0 & 0 & 0 & 1 & 0 & 0 & 0 \\
kilt\_wow & 0 & 0 & 0 & 0 & 0 & 0 & 1 & 0 & 0 & 0 & 0 & 1 & 0 & 0 & 0 & 0 & 1 & 0 & 0 & 0 \\
kilt\_zsre & 0 & 0 & 0 & 0 & 0 & 0 & 0 & 0 & 0 & 1 & 0 & 0 & 0 & 0 & 0 & 0 & 1 & 0 & 0 & 0 \\
lama-conceptnet & 0 & 0 & 0 & 0 & 0 & 0 & 0 & 0 & 0 & 0 & 1 & 0 & 0 & 0 & 0 & 0 & 1 & 0 & 1 & 0 \\
lama-google\_re & 0 & 0 & 0 & 0 & 0 & 0 & 0 & 0 & 0 & 0 & 1 & 0 & 0 & 0 & 0 & 0 & 1 & 0 & 1 & 0 \\
lama-squad & 0 & 0 & 0 & 1 & 0 & 0 & 0 & 0 & 0 & 0 & 1 & 0 & 0 & 0 & 0 & 0 & 1 & 0 & 1 & 0 \\
lama-trex & 0 & 0 & 0 & 0 & 0 & 0 & 0 & 0 & 0 & 0 & 1 & 0 & 0 & 0 & 0 & 0 & 1 & 0 & 1 & 0 \\
liar & 0 & 0 & 1 & 0 & 0 & 0 & 0 & 0 & 0 & 0 & 1 & 0 & 0 & 0 & 0 & 0 & 1 & 0 & 0 & 0 \\
limit & 0 & 0 & 0 & 0 & 0 & 0 & 0 & 0 & 0 & 1 & 0 & 0 & 1 & 0 & 0 & 1 & 0 & 0 & 0 & 0 \\
math\_qa & 1 & 0 & 0 & 0 & 0 & 0 & 0 & 0 & 0 & 0 & 0 & 1 & 1 & 0 & 1 & 0 & 0 & 0 & 0 & 0 \\
mc\_taco & 0 & 0 & 0 & 0 & 0 & 0 & 0 & 0 & 0 & 0 & 1 & 0 & 0 & 0 & 1 & 0 & 0 & 0 & 0 & 0 \\
medical\_questions\_pairs & 0 & 0 & 0 & 1 & 1 & 0 & 0 & 0 & 0 & 0 & 1 & 0 & 0 & 1 & 0 & 0 & 0 & 0 & 0 & 0 \\
mocha & 0 & 0 & 0 & 1 & 0 & 0 & 0 & 0 & 0 & 0 & 0 & 1 & 0 & 1 & 0 & 0 & 0 & 0 & 0 & 0 \\
multi\_news & 0 & 0 & 1 & 0 & 0 & 0 & 0 & 0 & 0 & 0 & 0 & 1 & 0 & 0 & 0 & 1 & 0 & 1 & 0 & 1 \\
numer\_sense & 0 & 0 & 0 & 1 & 0 & 0 & 0 & 0 & 0 & 0 & 1 & 0 & 0 & 0 & 1 & 0 & 1 & 0 & 1 & 0 \\
onestop\_english & 0 & 0 & 1 & 1 & 0 & 0 & 0 & 0 & 0 & 0 & 0 & 1 & 0 & 0 & 0 & 0 & 0 & 0 & 0 & 0 \\
openbookqa & 1 & 0 & 0 & 0 & 0 & 0 & 0 & 0 & 0 & 0 & 0 & 1 & 1 & 0 & 1 & 0 & 1 & 1 & 0 & 0 \\
paws & 0 & 0 & 0 & 1 & 0 & 0 & 0 & 0 & 0 & 0 & 1 & 0 & 0 & 1 & 0 & 0 & 0 & 0 & 0 & 0 \\
piqa & 1 & 0 & 0 & 0 & 0 & 0 & 0 & 0 & 0 & 0 & 0 & 1 & 0 & 0 & 1 & 0 & 0 & 0 & 0 & 0 \\
poem\_sentiment & 0 & 0 & 0 & 0 & 0 & 0 & 0 & 1 & 0 & 0 & 1 & 0 & 0 & 0 & 0 & 0 & 0 & 0 & 0 & 0 \\
proto\_qa & 0 & 0 & 0 & 0 & 0 & 0 & 0 & 0 & 0 & 0 & 1 & 0 & 0 & 0 & 1 & 0 & 0 & 0 & 0 & 0 \\
qa\_srl & 0 & 0 & 1 & 1 & 0 & 0 & 0 & 0 & 0 & 0 & 1 & 0 & 0 & 0 & 0 & 1 & 0 & 0 & 0 & 0 \\
qasc & 1 & 0 & 0 & 0 & 0 & 0 & 0 & 0 & 0 & 0 & 1 & 0 & 1 & 0 & 1 & 0 & 1 & 0 & 0 & 0 \\
quail & 0 & 0 & 1 & 1 & 0 & 0 & 0 & 1 & 0 & 0 & 0 & 1 & 1 & 0 & 0 & 0 & 0 & 0 & 0 & 0 \\
quarel & 0 & 0 & 0 & 0 & 0 & 0 & 0 & 0 & 0 & 0 & 1 & 0 & 1 & 0 & 1 & 0 & 0 & 0 & 1 & 0 \\
quartz-no\_knowledge & 0 & 0 & 0 & 0 & 0 & 0 & 0 & 0 & 0 & 0 & 1 & 0 & 1 & 0 & 1 & 0 & 0 & 0 & 1 & 0 \\
quartz-with\_knowledge & 0 & 0 & 0 & 0 & 0 & 0 & 0 & 0 & 0 & 0 & 1 & 0 & 1 & 0 & 0 & 0 & 0 & 0 & 1 & 0 \\
quoref & 0 & 0 & 0 & 1 & 0 & 0 & 0 & 0 & 0 & 0 & 0 & 1 & 0 & 0 & 0 & 1 & 0 & 0 & 0 & 0 \\
race-high & 0 & 0 & 0 & 0 & 0 & 0 & 0 & 0 & 0 & 0 & 0 & 1 & 1 & 0 & 0 & 0 & 0 & 0 & 1 & 0 \\
race-middle & 0 & 0 & 0 & 0 & 0 & 0 & 0 & 0 & 0 & 0 & 0 & 1 & 1 & 0 & 0 & 0 & 0 & 0 & 1 & 0 \\
reddit\_tifu-title & 0 & 1 & 0 & 1 & 0 & 0 & 0 & 0 & 0 & 0 & 0 & 1 & 0 & 0 & 0 & 1 & 0 & 0 & 0 & 1 \\
reddit\_tifu-tldr & 0 & 1 & 0 & 1 & 0 & 0 & 0 & 0 & 0 & 0 & 0 & 1 & 0 & 0 & 0 & 1 & 0 & 0 & 0 & 1 \\
ropes & 1 & 0 & 0 & 0 & 0 & 0 & 0 & 0 & 0 & 0 & 0 & 1 & 0 & 0 & 1 & 1 & 0 & 0 & 0 & 0 \\
rotten\_tomatoes & 0 & 0 & 0 & 0 & 0 & 1 & 0 & 0 & 0 & 0 & 1 & 0 & 0 & 0 & 0 & 0 & 0 & 0 & 0 & 0 \\
samsum & 0 & 0 & 0 & 0 & 0 & 0 & 1 & 0 & 0 & 0 & 0 & 1 & 0 & 0 & 0 & 1 & 0 & 0 & 0 & 1 \\
scicite & 1 & 0 & 0 & 0 & 0 & 0 & 0 & 0 & 0 & 0 & 1 & 0 & 0 & 0 & 0 & 0 & 0 & 0 & 0 & 0 \\
sciq & 1 & 0 & 0 & 0 & 0 & 0 & 0 & 0 & 0 & 0 & 0 & 1 & 1 & 0 & 0 & 0 & 0 & 0 & 0 & 0 \\
scitail & 1 & 0 & 0 & 0 & 0 & 0 & 0 & 0 & 0 & 0 & 1 & 0 & 0 & 0 & 0 & 0 & 0 & 0 & 0 & 0 \\
search\_qa & 0 & 0 & 0 & 0 & 0 & 0 & 0 & 0 & 0 & 0 & 1 & 0 & 0 & 0 & 0 & 0 & 1 & 0 & 0 & 0 \\
sick & 0 & 0 & 0 & 1 & 0 & 0 & 0 & 0 & 0 & 0 & 1 & 0 & 0 & 0 & 0 & 0 & 0 & 0 & 0 & 0 \\
sms\_spam & 0 & 1 & 0 & 0 & 0 & 0 & 0 & 0 & 0 & 0 & 1 & 0 & 0 & 0 & 0 & 0 & 0 & 0 & 0 & 0 \\
social\_i\_qa & 0 & 0 & 0 & 0 & 0 & 0 & 0 & 0 & 0 & 0 & 1 & 0 & 1 & 0 & 1 & 0 & 0 & 0 & 0 & 0 \\
spider & 0 & 0 & 0 & 0 & 0 & 0 & 0 & 0 & 0 & 0 & 1 & 0 & 0 & 0 & 0 & 0 & 0 & 0 & 0 & 1 \\
squad-no\_context & 0 & 0 & 0 & 0 & 0 & 0 & 0 & 0 & 0 & 0 & 1 & 0 & 0 & 0 & 0 & 0 & 1 & 0 & 0 & 0 \\
squad-with\_context & 0 & 0 & 0 & 1 & 0 & 0 & 0 & 0 & 0 & 0 & 0 & 1 & 0 & 0 & 0 & 1 & 0 & 0 & 0 & 0 \\
superglue-cb & 0 & 0 & 1 & 0 & 0 & 0 & 1 & 1 & 0 & 0 & 0 & 1 & 0 & 0 & 0 & 1 & 0 & 0 & 0 & 0 \\
superglue-copa & 0 & 1 & 0 & 1 & 0 & 0 & 0 & 0 & 0 & 0 & 1 & 0 & 1 & 0 & 1 & 0 & 0 & 0 & 0 & 0 \\
superglue-multirc & 1 & 0 & 1 & 1 & 0 & 0 & 0 & 1 & 0 & 0 & 0 & 1 & 1 & 0 & 0 & 0 & 0 & 0 & 0 & 0 \\
superglue-record & 0 & 0 & 1 & 0 & 0 & 0 & 0 & 0 & 0 & 0 & 0 & 1 & 0 & 0 & 1 & 1 & 0 & 0 & 0 & 0 \\
superglue-rte & 0 & 0 & 1 & 1 & 0 & 0 & 0 & 0 & 0 & 0 & 1 & 0 & 0 & 0 & 0 & 0 & 0 & 0 & 0 & 0 \\
superglue-wic & 0 & 0 & 0 & 0 & 0 & 0 & 0 & 0 & 0 & 0 & 1 & 0 & 0 & 1 & 1 & 0 & 0 & 0 & 0 & 0 \\
superglue-wsc & 0 & 0 & 0 & 0 & 0 & 0 & 0 & 1 & 0 & 0 & 0 & 1 & 0 & 0 & 0 & 1 & 0 & 0 & 0 & 0 \\
swag & 0 & 0 & 0 & 0 & 0 & 0 & 0 & 0 & 0 & 0 & 1 & 0 & 1 & 0 & 1 & 0 & 0 & 0 & 1 & 0 \\
tab\_fact & 0 & 0 & 0 & 1 & 0 & 0 & 0 & 0 & 0 & 0 & 0 & 1 & 0 & 0 & 0 & 0 & 0 & 1 & 0 & 0 \\
trec & 0 & 0 & 0 & 0 & 0 & 0 & 0 & 0 & 0 & 0 & 1 & 0 & 0 & 0 & 0 & 0 & 0 & 0 & 0 & 0 \\
trec-finegrained & 0 & 0 & 0 & 0 & 0 & 0 & 0 & 0 & 0 & 0 & 1 & 0 & 0 & 0 & 0 & 0 & 0 & 0 & 0 & 0 \\
tweet\_eval-emoji & 0 & 1 & 0 & 0 & 0 & 0 & 0 & 0 & 0 & 0 & 1 & 0 & 0 & 0 & 0 & 0 & 0 & 0 & 0 & 1 \\
tweet\_eval-emotion & 0 & 1 & 0 & 0 & 0 & 0 & 0 & 0 & 0 & 0 & 1 & 0 & 0 & 0 & 0 & 0 & 0 & 0 & 0 & 0 \\
tweet\_eval-hate & 0 & 1 & 0 & 0 & 0 & 0 & 0 & 0 & 0 & 0 & 1 & 0 & 0 & 0 & 0 & 0 & 0 & 0 & 0 & 0 \\
tweet\_eval-irony & 0 & 1 & 0 & 0 & 0 & 0 & 0 & 0 & 0 & 0 & 1 & 0 & 0 & 0 & 0 & 0 & 0 & 0 & 0 & 0 \\
tweet\_eval-offensive & 0 & 1 & 0 & 0 & 0 & 0 & 0 & 0 & 0 & 0 & 1 & 0 & 0 & 0 & 0 & 0 & 0 & 0 & 0 & 0 \\
tweet\_eval-sentiment & 0 & 1 & 0 & 0 & 0 & 0 & 0 & 0 & 0 & 0 & 1 & 0 & 0 & 0 & 0 & 0 & 0 & 0 & 0 & 0 \\
tweet\_eval-stance\_abortion & 0 & 1 & 0 & 0 & 0 & 0 & 0 & 0 & 0 & 0 & 1 & 0 & 0 & 0 & 0 & 0 & 0 & 0 & 0 & 0 \\
tweet\_eval-stance\_atheism & 0 & 1 & 0 & 0 & 0 & 0 & 0 & 0 & 0 & 0 & 1 & 0 & 0 & 0 & 0 & 0 & 0 & 0 & 0 & 0 \\
tweet\_eval-stance\_climate & 0 & 1 & 0 & 0 & 0 & 0 & 0 & 0 & 0 & 0 & 1 & 0 & 0 & 0 & 0 & 0 & 0 & 0 & 0 & 0 \\
tweet\_eval-stance\_feminist & 0 & 1 & 0 & 0 & 0 & 0 & 0 & 0 & 0 & 0 & 1 & 0 & 0 & 0 & 0 & 0 & 0 & 0 & 0 & 0 \\
tweet\_eval-stance\_hillary & 0 & 1 & 0 & 0 & 0 & 0 & 0 & 0 & 0 & 0 & 1 & 0 & 0 & 0 & 0 & 0 & 0 & 0 & 0 & 0 \\
tweet\_qa & 0 & 1 & 0 & 0 & 0 & 0 & 0 & 0 & 0 & 0 & 0 & 1 & 0 & 0 & 0 & 0 & 0 & 0 & 0 & 0 \\
web\_questions & 0 & 0 & 0 & 1 & 0 & 0 & 0 & 0 & 0 & 0 & 1 & 0 & 0 & 0 & 0 & 0 & 1 & 0 & 0 & 0 \\
wiki\_auto & 0 & 0 & 0 & 1 & 0 & 0 & 0 & 0 & 0 & 0 & 1 & 0 & 0 & 1 & 0 & 0 & 0 & 0 & 0 & 0 \\
wiki\_bio & 0 & 0 & 0 & 0 & 1 & 0 & 0 & 0 & 0 & 0 & 0 & 1 & 0 & 0 & 0 & 0 & 0 & 0 & 0 & 1 \\
wiki\_qa & 0 & 0 & 0 & 1 & 0 & 0 & 0 & 0 & 0 & 0 & 1 & 0 & 0 & 0 & 0 & 0 & 0 & 0 & 0 & 0 \\
wiki\_split & 0 & 0 & 0 & 1 & 0 & 0 & 0 & 0 & 0 & 0 & 1 & 0 & 0 & 1 & 0 & 1 & 0 & 0 & 0 & 1 \\
wikisql & 0 & 0 & 0 & 1 & 0 & 0 & 0 & 0 & 0 & 0 & 1 & 0 & 0 & 0 & 0 & 0 & 0 & 0 & 0 & 1 \\
wino\_grande & 0 & 0 & 0 & 0 & 0 & 0 & 0 & 0 & 0 & 1 & 0 & 0 & 1 & 1 & 1 & 1 & 0 & 0 & 1 & 0 \\
wiqa & 0 & 0 & 0 & 0 & 0 & 0 & 0 & 0 & 0 & 0 & 0 & 1 & 1 & 0 & 1 & 0 & 0 & 0 & 0 & 0 \\
xsum & 0 & 0 & 1 & 1 & 0 & 0 & 0 & 0 & 0 & 0 & 0 & 1 & 0 & 0 & 0 & 1 & 0 & 0 & 0 & 1 \\
yahoo\_answers\_topics & 1 & 1 & 0 & 1 & 0 & 0 & 0 & 0 & 0 & 0 & 0 & 1 & 0 & 0 & 0 & 0 & 0 & 0 & 0 & 0 \\
yelp\_polarity & 0 & 0 & 0 & 0 & 0 & 1 & 0 & 0 & 0 & 0 & 0 & 1 & 0 & 0 & 0 & 0 & 0 & 0 & 0 & 0 \\
yelp\_review\_full & 0 & 0 & 0 & 0 & 0 & 1 & 0 & 0 & 0 & 0 & 0 & 1 & 0 & 0 & 0 & 0 & 0 & 0 & 0 & 0 \\
       \Xhline{3\arrayrulewidth}
    \end{tabular}}
    \caption{Full feature table used for analysis in \S\ref{sec:interpret}.}
    \label{tab:hand_features}
\vspace{-7mm}
\end{table}

\normalsize
\newpage

\end{document}